%% file: main.tex
\documentclass{article}

\usepackage{microtype}
\usepackage{graphicx}
\usepackage{subcaption}
\usepackage{booktabs} 
\usepackage{titletoc}

\usepackage{hyperref}

\usepackage[preprint]{icml2026}

\usepackage{amsmath}
\usepackage{amssymb}
\usepackage{mathtools}
\usepackage{amsthm}

\usepackage[table,dvipsnames]{xcolor}

\usepackage{graphicx}
\usepackage{mathtools}
\usepackage{amsfonts}
\usepackage{booktabs}
\usepackage{color}
\usepackage{multirow}
\usepackage{subcaption}
\usepackage{caption}
\usepackage{makecell}
\usepackage{multicol}
\usepackage{pifont} 
\usepackage[table]{xcolor}
\definecolor{best}{RGB}{255,200,200}
\definecolor{secondbest}{RGB}{200,200,255}

\usepackage[capitalize,noabbrev]{cleveref}

\theoremstyle{plain}

\theoremstyle{definition}

\theoremstyle{remark}

\usepackage[textsize=tiny]{todonotes}

\icmltitlerunning{Towards Customized Multimodal Role-Play}

\begin{document}

\twocolumn[
  \icmltitle{Towards Customized Multimodal Role-Play}



  \icmlsetsymbol{equal}{*}

  \begin{icmlauthorlist}
    \icmlauthor{Chao Tang}{sch,comp}
    \icmlauthor{Jianzong Wu}{sch,comp}
    \icmlauthor{Qingyu Shi}{sch}
    \icmlauthor{Ye Tian}{sch}
    \icmlauthor{Aixi Zhang}{comp}
    \\
    \icmlauthor{Hao Jiang}{comp}
    \icmlauthor{Jiangning Zhang}{yyy}
    \icmlauthor{Yunhai Tong}{sch}
  \end{icmlauthorlist}

  \icmlaffiliation{sch}{Peking University}
  \icmlaffiliation{comp}{Alibaba Group}
  \icmlaffiliation{yyy}{Zhejiang University}

  \icmlcorrespondingauthor{Chao Tang}{tc\_2003@stu.pku.edu.cn}

  \icmlkeywords{Machine Learning, ICML}

  {%
\renewcommand\twocolumn[1][]{#1}%
\begin{center}
    \centering
    \captionsetup{type=figure}
    \includegraphics[width=1\textwidth]{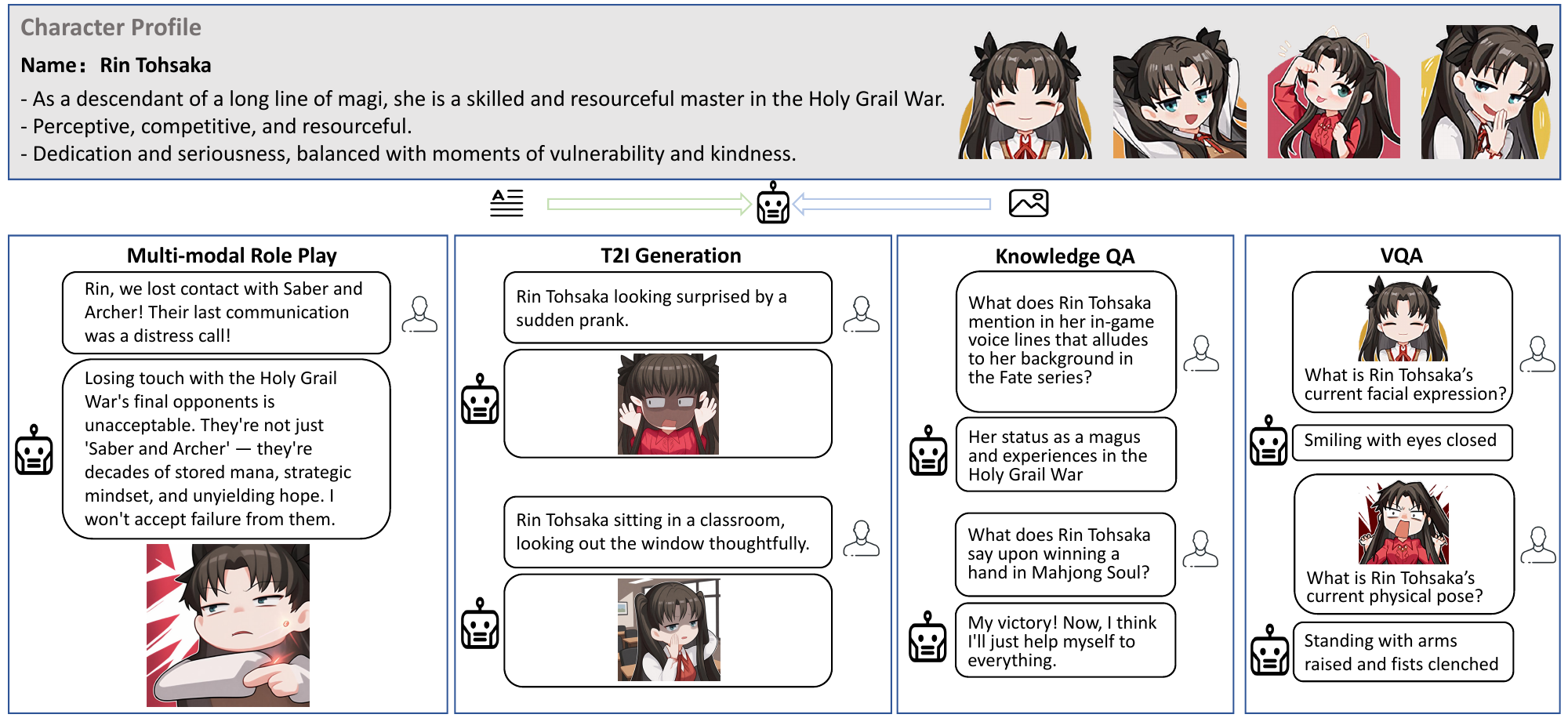}
    \vspace{-0.5cm}
    \captionof{figure}{\textbf{Demonstration of the UniCharacter model's capabilities.} The model utilizes a character's profile to maintain consistency across several integrated tasks. The core innovation is showcased in Multimodal Role Play, where the model simultaneously generates a coherent textual response and a corresponding visual image that reflects the character's emotion. This unified generation is supplemented by the model's ability to perform Text-to-Image (T2I) Generation, Knowledge Question-Answering (Knowledge QA), and Visual Question-Answering (VQA). Together, these functions highlight UniCharacter's ability to create a cohesive, interactive, and visually embodied persona within a single framework.}
    \label{fig:teaser}
\end{center}%
}

  \vskip 0.3in
]



\printAffiliationsAndNotice{}  

\input{sec/0_abstract}    
\input{sec/1_intro}
\input{sec/2_related_work}
\input{sec/3_dataset}
\input{sec/4_method}
\input{sec/5_exp}
\input{sec/6_conclusion}

\section*{Acknowledgements}

This work is supported by the National Key Research and Development Program of China (No. 2023YFC3807600).

\section*{Impact Statement}




This paper introduces UniCharacter, a framework for Customized Multimodal Role-Play that leverages the capabilities of unified multimodal models to create interactive agents that are both character-rich and deeply immersive. While our work carries broader societal implications, we believe there are no specific concerns that necessitate detailed discussion in this context.

\nocite{langley00}

\bibliography{example_paper}
\bibliographystyle{icml2026}


\newpage

\input{sec/X_suppl}

\end{document}

%% file: sec/0_abstract.tex
\begin{abstract}
Unified multimodal understanding and generation models enable richer human-AI interaction. 
Yet jointly customizing a character's persona, dialogue style, and visual identity while maintaining output consistency across modalities remains largely unexplored. 
To mitigate this gap, we introduce a new task, Customized Multimodal Role-Play (CMRP).
We construct the \textbf{RoleScape-20 dataset} comprising 20 characters, including training and evaluation data that cover persona, stylistic descriptions, visual/expressive cues, and text–image interactions.
%
Building on a unified model, we devise \textbf{UniCharacter}, a two-stage training framework containing Unified Supervised Finetuning (Unified-SFT) and character-specific group relative policy optimization (Character-GRPO). 
Given only 10 images plus corresponding interaction examples, the model acquires the target character and exhibits coherent persona, style, and visual identity in both generated text and images. This process takes about 100 GPU hours.
Experiments on the RoleScape-20 dataset show that the proposed method substantially outperforms prior approaches. 
Ablation studies further validate the effectiveness of our cross-modal consistency design and few-shot customization strategy. 
We argue that CMRP, coupled with unified modeling, provides a basis for next-generation characterful and immersive interactive agents.
Our dataset and code will be released at:~\href{https://github.com/Tangc03/UniCharacter}{https://github.com/Tangc03/UniCharacter}
\end{abstract}

%% file: sec/1_intro.tex
\section{Introduction}
\label{sec:intro}
Personalized virtual characters are increasingly used in digital avatars, interactive entertainment, and human–AI communication.
Existing systems usually operate in a \textit{single modality}. Text-based models~\citep{rolellm,characterllm, yollava} can be customized for persona-aligned role-play but cannot generate visual content. 
Image personalization methods~\citep{dreambooth,textual_inversion, jedi} can reproduce a character’s appearance, but cannot participate in conversations or react to contextual cues. Current approaches can only customize how a character speaks or looks, but not both at the same time.

Recent unified multimodal foundation models~\citep{januspro, bagel, mmada, showo} offer a promising way to bridge this gap. These models process and generate both text and images within a single architecture, and they already demonstrate strong cross-modal understanding and generation capabilities. 
They could support virtual characters that are both linguistically expressive and visually creative. 
Yet, current cases of these models~\citep{unictokens, yochameleon} focus on tasks such as Visual Question Answering (VQA), captioning, or general text–to-image generation. 
None of them targets persona-driven interaction that requires consistent language style, emotional expression, and visual identity. Neglecting identity consistency across modalities prevents models from effectively constructing a complete multimodal character. This is also a crucial foundation for achieving more immersive user-character interactions in role-playing scenarios, holding significant application potential.

To address this gap, we introduce \textbf{Customized Multimodal Role-Play (CMRP)}, a task that adapts a general-purpose multimodal model into a virtual character using minimal character-specific data—a textual profile, a few reference images, and example dialogues—to generate in-character responses and appearance-consistent images for interactive role-play with a stable persona and visual identity.

To facilitate CMRP, we introduce RoleScape-20, the first multimodal role-play dataset. 
It has 20 diverse characters, each with a textual profile, 5--15 reference images, and 150--250 role-playing dialogues. 
We also provide fine-grained multimodal annotations, including explicit thinking processes, image-generation instructions, and paired visual or knowledge-based QA samples. 
These components support unified modeling of persona, language, and visual identity.

Building on this dataset, we propose \textbf{UniCharacter}, a framework that adapts unified multimodal models to coherent multimodal role-play via a two-stage pipeline. 
Stage 1 performs Unified Supervised Fine-Tuning (Unified-SFT) across all tasks. 
However, image generation SFT relies on ground-truth images, which limits scaling and leads to overfitting and low output diversity. 
Therefore, Stage 2 introduces Character Group Relative Policy Optimization (Character-GRPO) for text-to-image (T2I) generation, hoping that its group-based sampling pipeline can encourage the model to explore diverse visual representations, and its training data requirement, with no ground truth images needed, can further expand the diversity of image generation scenarios.
%
We apply GRPO training to the T2I generation task. 
By using rewards for text-image alignment, group diversity, and a penalty for similarity to training images, our Character-GRPO training stage effectively enhances the diversity of model outputs in image generation tasks.

Extensive experiments demonstrate that UniCharacter surpasses competitive baselines (e.g., UniCTokens~\cite{unictokens}, DreamBooth~\cite{dreambooth}, Qwen2.5-VL~\cite{qwen2_5-vl})
in role consistency, dialogue authenticity, image fidelity, and cross-modal alignment, advancing the creation of coherent, lifelike virtual agents.

\cref{tab:method-compare} summarizes the differences between UniCharacter and recent works, and our contributions are as follows:
\begin{itemize}
    \item We introduce CMRP, a new task for multimodal role-play that integrates both textual role-playing and multimodal personalization, along with RoleScape-20, the first dataset designed for multimodal role-play.
    \item We propose UniCharacter, a two-stage framework comprising Unified-SFT and Character-GRPO, for few-shot vision-language alignment. Character-GRPO employs a reward mechanism to mitigate T2I overfitting while preserving text-image consistency.
    \item Extensive experiments show that our approach outperforms baselines in role consistency, dialogue quality, image fidelity, and cross-modal alignment.
\end{itemize}

\begin{table}[t]
    \centering
    \caption{\textbf{Comparison between UniCharacter and recent works.}
    }
    \label{tab:method-compare}
    \scalebox{0.45}{
    \begin{tabular}{cccccc}
        \toprule
        Method & Text Role-Play & Multimodal Role-Play & T2I Generation & Knowledge QA & VQA \\
        \midrule
        CharacterLLM~\cite{characterllm} & \color{blue}\ding{51} & \color{red}\ding{55} & \color{red}\ding{55} & \color{blue}\ding{51} & \color{red}\ding{55} \\
        DreamBooth~\cite{dreambooth} & \color{red}\ding{55} & \color{red}\ding{55} & \color{blue}\ding{51} & \color{red}\ding{55} & \color{red}\ding{55} \\
        \midrule
        Yo'LLaVA~\cite{yollava} & \color{red}\ding{55} & \color{red}\ding{55} & \color{red}\ding{55} & \color{red}\ding{55} & \color{blue}\ding{51} \\
        MyVLM~\cite{myvlm} & \color{red}\ding{55} & \color{red}\ding{55} & \color{red}\ding{55} & \color{red}\ding{55} & \color{blue}\ding{51} \\
        \midrule
        Yo'Chamaleon~\cite{yochameleon} & \color{red}\ding{55} & \color{red}\ding{55} & \color{blue}\ding{51} & \color{red}\ding{55} & \color{blue}\ding{51} \\
        UniCTokens~\cite{unictokens} & \color{red}\ding{55} & \color{red}\ding{55} & \color{blue}\ding{51} & \color{red}\ding{55} & \color{blue}\ding{51} \\
        \midrule
        \textbf{UniCharacter (Ours)} & \textbf{\color{blue}\ding{51}} & \textbf{\color{blue}\ding{51}} & \textbf{\color{blue}\ding{51}} & \textbf{\color{blue}\ding{51}} & \textbf{\color{blue}\ding{51}} \\
        \bottomrule
    \end{tabular}}
    \vskip -0.2in
\end{table}

%% file: sec/2_related_work.tex
\section{Related Work}
\label{sec:related_work}

\begin{table*}[t]
    \centering
    \caption{\textbf{Comparison between RoleScape-20 and related datasets.} \textit{Multimodal Role-Play Data} denotes paired visual-textual role-playing episodes where images depict the character in a context that aligns with their dialogue, emotional state, and personality. \textit{Char} refers to character. \textit{Img} refers to Image.}
    \label{tab:dataset-compare}
    \scalebox{0.60}{
    \begin{tabular}{lcccccc|cccccc}  
\toprule
\multirow{4}{*}{Dataset} & \multirow{4}{*}{Modality} & \multirow{4}{*}{\#Chars} & \multirow{4}{*}{\#Img/Char} & \multirow{4}{*}{\#Dialogues/Char} & \multirow{4}{*}{\#VQA/Char} & \multirow{4}{*}{\#QA/Char} & \multicolumn{6}{c}{Annotations} \\
\cmidrule{8-13}
& & & & & & & \makecell{Role-Play\\Dialogue} &  \makecell{Knowledge\\QA} & VQA & \makecell{Character\\Images} & \makecell{Multimodal\\Role-Play\\Data} & \makecell{Thinking\\Process} \\
\midrule
CharacterLLM~\cite{characterllm} & Text & 9 & – & 1.6K & - & - & \color{blue}\ding{51} & \color{red}\ding{55} & \color{red}\ding{55} & \color{red}\ding{55} & \color{red}\ding{55} & \color{red}\ding{55} \\
ChatHaruhi-54K~\cite{chatharuhi} & Text & 32 & – & 1.7K & - & - & \color{blue}\ding{51} & \color{red}\ding{55} & \color{red}\ding{55} & \color{red}\ding{55} & \color{red}\ding{55} & \color{red}\ding{55}  \\
\midrule
DreamBooth~\cite{dreambooth} & Image & 30 & 3–5 & - & - & - & \color{red}\ding{55} & \color{red}\ding{55} & \color{red}\ding{55} & \color{blue}\ding{51} & \color{red}\ding{55} & \color{red}\ding{55} \\
\midrule
Yo'LLaVA~\cite{yollava} & Image+Text & 40 & 5-10 & - & \textasciitilde4 & - & \color{red}\ding{55} & \color{red}\ding{55}  & \color{blue}\ding{51} & \color{blue}\ding{51} & \color{red}\ding{55} & \color{red}\ding{55}  \\
MyVLM~\cite{myvlm} & Image+Text & 30 & 10 & - & 90 & - & \color{red}\ding{55} & \color{red}\ding{55} & \color{blue}\ding{51} & \color{blue}\ding{51} & \color{red}\ding{55} & \color{red}\ding{55}  \\
UnifyBench~\cite{unictokens} & Image+Text & 20 & 5-10 & - & \textasciitilde200 & - & \color{red}\ding{55} & \color{red}\ding{55} & \color{blue}\ding{51} & \color{blue}\ding{51} & \color{red}\ding{55} & \color{red}\ding{55} \\
\midrule
\textbf{RoleScape-20 (Ours)} & \textbf{Image+Text} & \textbf{20} & \textbf{5-15} & \textbf{150-250} & \textbf{\textasciitilde200} & \textbf{\textasciitilde100} & \textbf{\color{blue}\ding{51}} & \textbf{\color{blue}\ding{51}} & \textbf{\color{blue}\ding{51}} & \textbf{\color{blue}\ding{51}} & \textbf{\color{blue}\ding{51}} & \textbf{\color{blue}\ding{51}} \\
\bottomrule
    \end{tabular}}
\vskip -0.1in
\end{table*}

%
\noindent
\textbf{Customized Generation.}
Customized generation generates content that follows the role specified by the user, represented through text~\cite{character,coser,personavectors} or images~\cite{textual_inversion,animatediff,custom_diffusion,DiffSensei,LGVI,CGG}. Previous customization methods can be broadly divided into two categories: training-based~\cite{ip_adapter,jedi} and tuning-based~\cite{rolellm,characterllm,dreambooth,DreamRelation} approaches. Training-based methods introduce an extra module to encode the user input and guide the generation process. 
Tuning-based methods, on the other hand, finetune part of the model parameters to learn the user-provided role. 
They use a special token during finetuning and insert it at inference time for customization, achieving strong role fidelity and controllability.

However, existing methods~\cite{chatharuhi,yollava,mcllava} are limited to a single modality. During inference, the model customizes through text or images only~\cite{myvlm,repic,rap,DeT}, making it difficult to support interactions requiring both outputs. 
To address this, we propose Customized Multimodal Role-Play, a new task requiring joint text and image generation from user inputs, and introduce UniCharacter, a tuning-based method for this setting.

\noindent
\textbf{Customized Unified Multimodal Models.}
Unified models now integrate multimodal understanding and generation~\cite{bagel,omnigen2,showo2,januspro,emu3_5,muddit,mmada}, yet coherent personalization remains challenging. While Yo’Chameleon~\cite{yochameleon} uses disjoint strategies and UniCTokens~\cite{unictokens} tackles unified personalization, neither supports complex interactive scenarios. We address this with a framework that jointly models key dimensions, including persona, dialogue style, visual identity, and emotion.
Our approach extends unified models to complex personalized interactions and enables agents with consistent personalities and cross-modal coherence.
Meanwhile, Group Relative Policy Optimization (GRPO) has gained traction as an RL-based tuning method since DeepSeek-R1~\cite{deepseek-r1}, with work adapting it to unified models~\cite{ulm-r1, unirl} or flow-matching image generators~\cite{flow-grpo, diffusionnft}.
These works are based on general tasks and datasets, so we bridge these directions by incorporating GRPO into the rectified flow-based image generation branch of unified multimodal models, using a tailored reward function for CMRP to increase generation diversity while maintaining image quality and text-image alignment.

%% file: sec/3_dataset.tex
\section{The RoleScape-20 Dataset}
\label{sec:dataset}

\begin{figure*}[t!]
	\centering
	\includegraphics[width=1.0\linewidth]{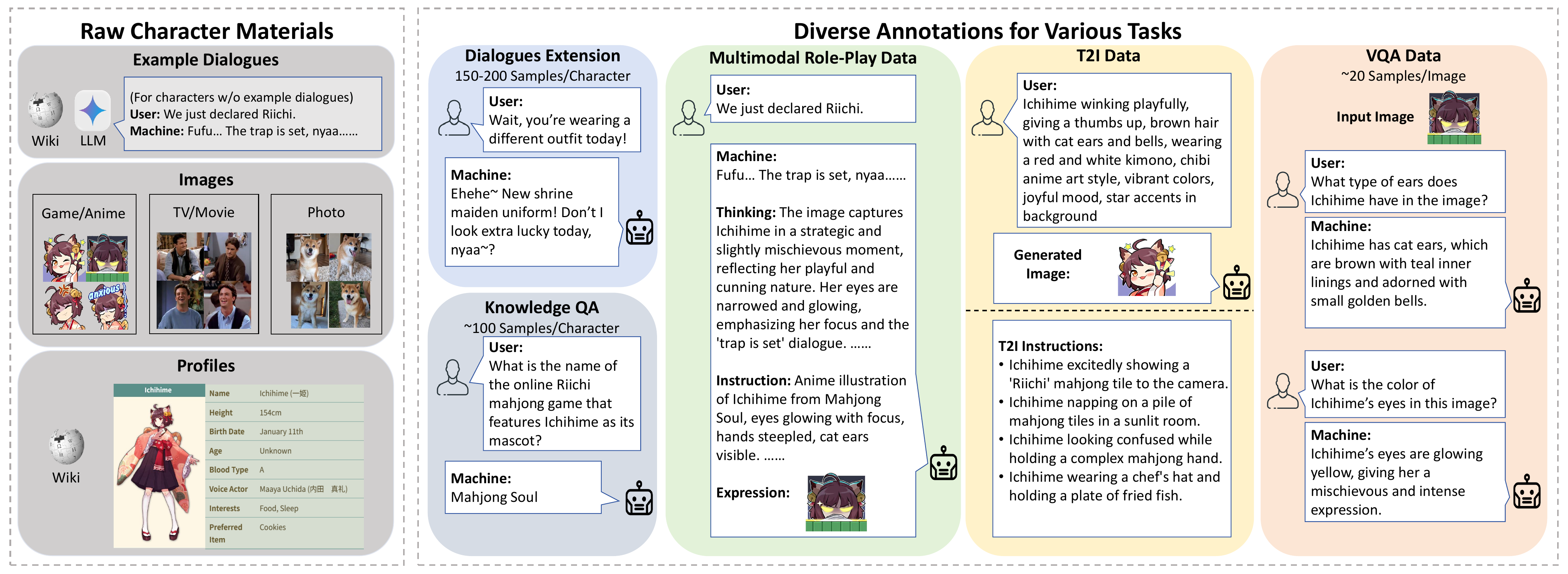}
	\caption{\textbf{Data Construction Pipeline of RoleScape-20 Dataset.} The data construction pipeline processes raw character materials (dialogues, images, profiles) into diverse training data, including multimodal role-play dialogues, T2I generation pairs, knowledge QA, and VQA pairs.}
	\label{fig:data-pipeline}
    \vskip -0.2in
\end{figure*}

\subsection{Problem Formulation}
We define the core task as Customized Multimodal Role-Play (CMRP), which aims to develop a computational agent that can faithfully emulate a specific virtual character based on a comprehensive character definition. A specific character is defined by a triplet $\mathcal{C} = \{\mathcal{P}_{\text{char}}, \mathcal{I}_{\text{core}}, \mathcal{D}_{\text{ref}}\}$. $\mathcal{P}_{\text{char}}$ is the textual profile, describing the character's personality, background, and traits. $\mathcal{I}_{\text{core}}$ is a set of core reference images that define the character's visual identity. $\mathcal{D}_{\text{ref}}$ is a collection of reference dialogues capturing the character's unique speaking style and linguistic habits.

Given the character definition $\mathcal{C}$ and a user's textual query $Q_u$, the CMRP task requires the model $\mathcal{F}_{\theta}$ to generate a multimodal response pair $(R_m, I_m)$, 
which must satisfy two key constraints: $R_m$ must follow the personality and speaking style defined in $\mathcal{P}_{\text{char}}$ and $\mathcal{D}_{\text{ref}}$ and $I_m$ must accurately depict the character's visual features as specified in $\mathcal{I}_{\text{core}}$ while being contextually relevant to $R_m$ and $Q_u$.
Formally, this interaction is represented as: $(R_m, I_m) = \mathcal{F}_{\theta}(Q_u)$.

Conceptually, this output is a sample from a conditional joint probability distribution shaped by the character dataset: $(R_m, I_m) \sim P(R, I | Q_u, \mathcal{C}; \theta)$. This joint probability can be decomposed via chain rule into two sequential stages: text generation $R_m \sim P(R | Q_u, \mathcal{C}; \theta)$ followed by conditional image generation $I_m \sim P(I | R_m, Q_u, \mathcal{C}; \theta)$.

The model integrates four core CMRP capabilities.

\noindent
\textbf{Multimodal Role-Play.} This is the primary task where the model acts as the character. Given a user query $Q_u$, the model generates a text-image response pair: $(R_m, I_m) \sim P(R, I \mid Q_u, \mathcal{C}; \theta)$. The response must maintain high persona consistency in linguistic style and visual identity.

\noindent
\textbf{Text-to-Image (T2I) Generation.} This capability focuses on the model's ability to translate a textual instruction or scene description $C_j$ into a high-quality consistent image $I_j$ that adheres to the character's visual identity in $\mathcal{I}_{\text{core}}$, modeled as: $I_j \sim P(I \mid C_j, \mathcal{C}; \theta)$.

\noindent
\textbf{Visual Question Answering (VQA).} VQA evaluates the model's understanding of the character's visual attributes. Given a reference image $I_{\text{ref}} \in \mathcal{I}_{\text{core}}$ and a specific question $Q_v$ regarding its details, the model must provide an accurate textual answer $A_v$: $A_v \sim P(A \mid I_{\text{ref}}, Q_v, \mathcal{C}; \theta)$.

\noindent
\textbf{Knowledge Question Answering (Knowledge QA).} This task requires the model to recall and reason over the character's background information. Given a textual question $Q_k$ about the character's life, traits, or history, the model retrieves the answer $A_k$ from $\mathcal{P}_{\text{char}}$: $A_k \sim P(A \mid Q_k, \mathcal{C}; \theta)$.

\subsection{Dataset Construction}

\noindent
\textbf{Overview.}
Existing general-purpose image-text dialogue datasets are insufficient for the demands of deep character customization. To address this, we construct RoleScape-20, a new dataset specifically designed for the CMRP task. It comprises 20 diverse characters organized into three main categories: nine real-world figures, mostly from movies and TV series, seven anime and game characters, and four animals. The raw materials for our dataset are sourced from various channels to ensure richness and authenticity. Images are collected from real photographs and high-resolution screenshots from films, television shows, games, and anime. Dialogues are compiled from authentic conversations found online and further supplemented and stylized using Large Language Models (LLMs) to align with character personas. Character profiles are sourced from authoritative sources like Wikipedia for real figures or generated by LLMs based on established lore for fictional characters. All materials undergo manual inspection and screening.

\begin{figure*}[t!]
	\centering
	\includegraphics[width=1.0\linewidth]{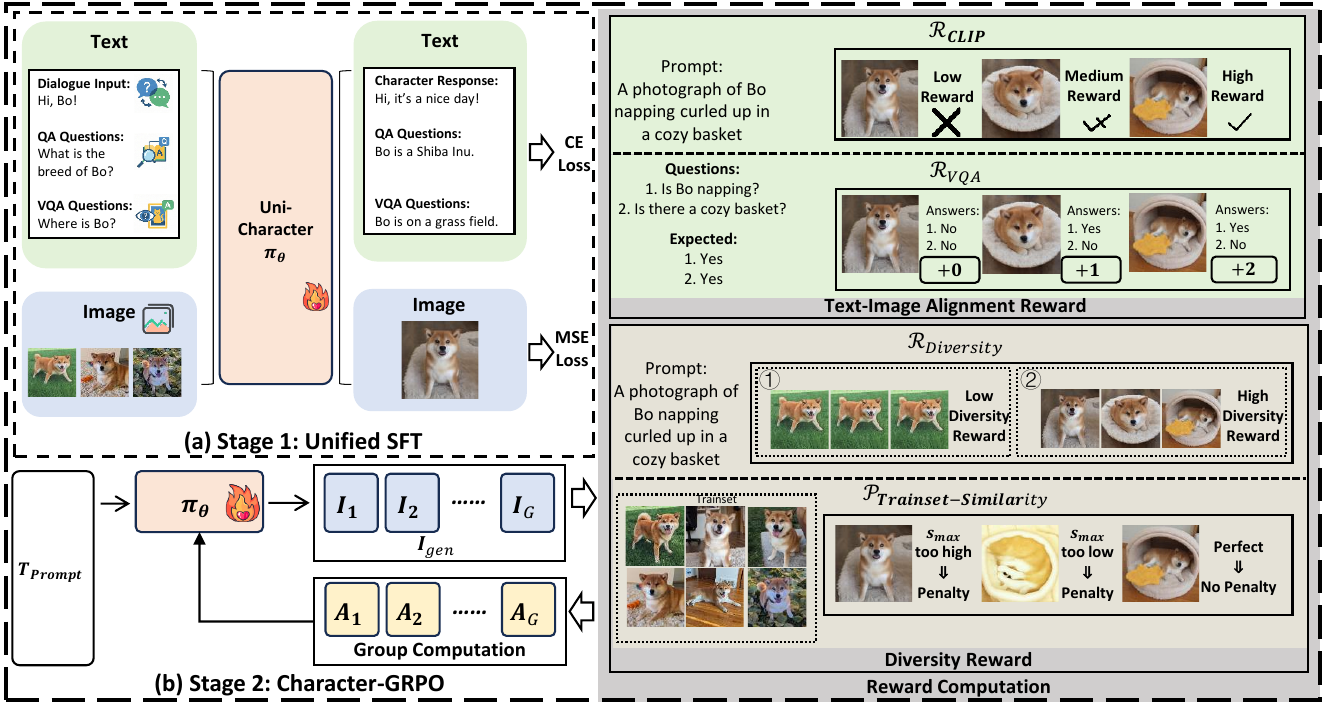}
	\caption{\textbf{Overview of the UniCharacter framework.} Stage 1 focuses on Unified-SFT, using MSE loss for image outputs and CE loss for text outputs. Stage 2 implements Character-GRPO, optimizing the policy $\pi_{\theta}$ via a multi-reward mechanism that considers both text-image alignment and generation diversity. }
	\label{fig:model-pipeline}
    \vskip -0.1in
\end{figure*}

\noindent
\textbf{Comparison with Related Datasets.}
RoleScape-20 fills a critical gap in the existing landscape of role-playing and customization datasets. Compared to text-only role-playing datasets like Character-LLM~\cite{characterllm}, and ChatHaruhi-54K~\cite{chatharuhi}, our dataset introduces the essential visual modality required for training and evaluating multimodal consistency. In contrast to personalized multimodal datasets such as Yo’LLaVA~\cite{yollava}, MyVLM~\cite{myvlm}, and UnifyBench~\cite{unictokens}, which often lack deep, in-character conversations, RoleScape-20 provides rich, personality-driven dialogues instead of simple image descriptions. Furthermore, unlike image customization datasets such as DreamBooth~\cite{dreambooth}, which focus narrowly on visual generation from simple, standardized captions, our dataset provides complex, multifaceted textual annotations, including conversational context, reasoning processes, and character knowledge. RoleScape-20 is the first dataset to provide this comprehensive suite of annotations, including fine-grained generation instructions, thinking processes, and both knowledge-based and visual question-answering pairs, establishing a solid foundation for training truly deep and consistent multimodal role-playing models. A more detailed comparison with previous datasets is presented in~\cref{tab:dataset-compare}.

\noindent
\textbf{Construction Pipeline.}
We designed a systematic annotation pipeline, shown in \cref{fig:data-pipeline}, to process the raw materials into a multi-faceted dataset capable of comprehensively training all the required model capabilities. The process consists of four main stages. First, to extend the amount of dialogues, we use Qwen3~\cite{qwen3} LLM to expand upon the initial reference dialogues ($\mathcal{D}_{\text{ref}}$) and character profile ($\mathcal{P}_{\text{char}}$), generating 150-200 new dialogue samples that faithfully mimic the character and produce the final text-only dialogue set, $\mathcal{D}_{\text{dialogue}}$. Second, for multimodal role-play and T2I generation data annotation, we take a core image ($I_j$) and its corresponding dialogue pair $(Q_u, R_m)$ as input. Using GPT-4o~\cite{gpt4o}, we generate two crucial annotations: a ``Thinking Process" that guides the image generation process, and a clear ``Instruction" to guide image generation for that specific context, resulting in a richly annotated data tuple $(I_j, Q_u, R_m, \text{Thinking Process}, \text{Instruction})$. Third, to construct the knowledge QA Dataset ($\mathcal{D}_{\text{kqa}}$), we employ an LLM to extract key information from the character profile ($\mathcal{P}_{\text{char}}$) and automatically convert it into question-answer pairs. Finally, for the visual QA Dataset ($\mathcal{D}_{\text{vqa}}$), we use the Qwen3-VL~\cite{qwen3,qwen-vl,qwen2-vl,qwen2_5-vl} multimodal model, providing it with a core image ($I_j$) and the character profile for context, to generate approximately 20 question-answer pairs focused on specific visual details within the image. All annotations are manually verified to confirm their authenticity. Data construction details are in the Appendix~\cref{sec:supp-detailed-data-construction-pipeline}.


%% file: sec/4_method.tex
\section{Method}
\label{sec:method}

An overview of UniCharacter is shown in \cref{fig:model-pipeline}. Detailed preliminaries are in the Appendix~\cref{sec:supp-preliminaries}.

\begin{table*}[!t]
\centering
\caption{\textbf{Comparison with previous works on RoleScape-20 benchmark.} TP = Text Prompt. Higher values are better. \textcolor{red}{Best} and \textcolor{blue}{Second best} performances are highlighted.}
\label{tab:exp-ours-vs-baselines}
\scalebox{0.66}{
\begin{tabular}{l|c|ccc|ccc|ccc|c|c}
\toprule
\multirow{2}{*}{Method} & \multirow{2}{*}{\makecell{Model\\Size}} & \multicolumn{3}{c|}{Text-based Role Play} & \multicolumn{3}{c|}{T2I} & \multicolumn{3}{c|}{Multimodal Role-Play (T2T2I)} & \multirow{2}{*}{Knowledge QA} & \multirow{2}{*}{VQA} \\
\cmidrule(lr){3-5} \cmidrule(lr){6-8} \cmidrule(lr){9-11}
 & & Memorization & Personality & Diversity & CLIP-I & CLIP-T & DINO & CLIP-I & CLIP-T & DINO & & \\
\midrule
DreamBooth~\cite{dreambooth} & 7B & - & - & - & \cellcolor{secondbest}0.86 & 0.30 & \cellcolor{secondbest}0.88 & - & - & - & - & - \\
Qwen2.5-VL+TP~\cite{qwen2_5-vl} & 7B & \cellcolor{secondbest}{5.13} & \cellcolor{secondbest}5.17 & \cellcolor{secondbest}{5.60} & - & - & - & - & - & - & \cellcolor{secondbest}{0.75} & \cellcolor{secondbest}{0.81} \\
UniCTokens~\cite{unictokens} & 1.3B & 2.43 & 2.54 & 2.30 & 0.73 & \cellcolor{best}0.34 & 0.83 & 0.51 & 0.17 & 0.70 & 0.08 & 0.21 \\
\midrule
UniCharacter(Ours) & 7B & \cellcolor{best}5.45  & \cellcolor{best}{6.55}  & \cellcolor{best}{6.10}  & \cellcolor{best}{0.88}  & \cellcolor{secondbest}{0.33}  & \cellcolor{best}{0.91}  & \cellcolor{best}{0.86}  & \cellcolor{best}{0.33}  & \cellcolor{best}{0.89}  & \cellcolor{best}0.77  & \cellcolor{best}0.84  \\
\bottomrule
\end{tabular}}
\vskip -0.1in
\end{table*}

\subsection{Unified Supervised Finetuning}
\label{subsection:unified-sft}

We frame the unified-SFT process as a multi-task learning problem, which is divided into two task categories.

\noindent
\textbf{Finetuning for Text Generation.}
This category of tasks, which we refer to as Vision-Language Understanding tasks, is designed to enhance the model's ability to comprehend and express the character's non-visual attributes. It includes four sub-tasks: Role-Play Chatting, where the model learns to generate in-character text responses; Thinking Task, where it learns to generate a reasoning process of character image generation; Visual Question Answering (VQA), where it answers questions based on a character image; and Knowledge Question Answering (Knowledge QA), where it answers questions about the character. The optimization objective for each of these tasks is to maximize the conditional probability of the target text, and the loss for each ($\mathcal{L}_{\text{chat}}, \mathcal{L}_{\text{think}}, \mathcal{L}_{\text{vqa}}, \mathcal{L}_{\text{kqa}}$) is calculated using a standard Cross-Entropy (CE) Loss. The total loss for this category is a weighted sum of the individual task losses:
\begin{equation}
\mathcal{L}_{\text{VLM}} = \mathcal{L}_{\text{chat}} + \mathcal{L}_{\text{think}} + \mathcal{L}_{\text{vqa}} + \mathcal{L}_{\text{kqa}}
\end{equation}

\noindent
\textbf{Finetuning for Image Generation.}
The T2I generation task generates image $I_{\text{gen}}$ based on text $C$. We use a Rectified Flow-based approach, where the loss function is the mean squared error (MSE) on the noise-to-clean residual.


\subsection{Character-GRPO}
\label{subsection:character-grpo}
Although the Unified SFT stage enables the model to perform well in textual dialogue, the Text-to-Image (T2I) branch often suffers from visual overfitting, resulting in generated images that lack variety. 
This is primarily because SFT relies on a limited set of fixed ground-truth images. 
To address this, we introduce \textbf{Character-GRPO}, a reinforcement learning stage dedicated to the T2I branch. 
In this stage, the model is no longer restricted to a single ground-truth image; instead, it generates multiple samples $\{I_1, I_2, \dots, I_G\}$ for each character-specific prompt. 
This multi-sample generation allows the model to explore a broader generation space, effectively mitigating overfitting. 
Furthermore, since GRPO does not require ground-truth images, it serves as a self-evolving data expansion mechanism that increases the diversity and volume of character-image mappings beyond the original training set.
To guide the policy $\pi_\theta$ toward generating character-consistent and diverse multimodal content, we define a comprehensive reward function $R_{total}$ comprising alignment and diversity components.

\noindent
\textbf{Text-Image Alignment Rewards. }
These rewards, shown in~\cref{fig:model-pipeline}, ensure that the generated visual content $I_{gen}$ adheres to the textual prompt $T_{prompt}$ and the character's intrinsic attributes.
The CLIP Similarity Reward ($r_{CLIP}$) measures the semantic alignment between the image and the prompt using CLIP:
\begin{equation}
    r_{CLIP} = \cos(\phi_I(I_{gen}), \phi_T(T_{prompt}))
\end{equation}
where $\phi_I$ and $\phi_T$ denote the CLIP image and text encoders, respectively.
The VQA Consistency Reward ($r_{VQA}$) verifies fine-grained character traits based on the correctness of the model's answers to specific annotated questions:
\begin{equation}
    r_{VQA} = 
    \begin{cases} 
        1, & \text{if VQA}(I_{gen}, T_{VQA}) \text{ matches } T_{truth} \\ 
        0, & \text{otherwise} 
    \end{cases}
\end{equation}

\noindent
\textbf{Diversity Rewards.}
To avoid overfitting and encourage diverse generation, we penalize redundancy and similarity to the training set, shown in~\cref{fig:model-pipeline}.
The Perceptual Diversity Reward ($r_{div}$) utilizes the Learned Perceptual Image Patch Similarity (LPIPS) to measure the visual variance within the sampled group:
\begin{equation}
    r_{div} = \frac{1}{G(G-1)} \sum_{i=1}^{G} \sum_{j \neq i}^{G} \text{LPIPS}(I_i, I_j)
\end{equation}
The Trainset Similarity Penalty ($p_{sim}$) prevents the model from memorizing training samples while ensuring it retains the target character's essential features. $p_{sim}$ is calculated with an upper threshold $\tau_{high}$ and a lower threshold $\tau_{low}$:
\begin{equation}
    p_{sim} =
    \begin{cases}
        - (s_{max} - \tau_{high}), & \text{if } s_{max} > \tau_{high} \\
        - (\tau_{low} - s_{max}), & \text{if } s_{max} < \tau_{low} \\
        0, & \text{otherwise}
    \end{cases}
\end{equation}
where $s_{max}$ refers to maximum cosine similarity between $I_{gen}$ and the training set $\mathcal{D}_{train}$ in the DINO feature space:
$$
    s_{max} = \max_{I_k \in \mathcal{D}_{train}} \left( \text{Sim}_{\text{DINO}}(I_{gen}, I_k) \right)
$$

\noindent
\textbf{Comprehensive Reward. }
The final reward $R_i$ for a sample in the group is a combination of the above components, providing a signal for GRPO's advantage computation:
\begin{equation}
    R_i = \underbrace{\alpha \cdot r_{CLIP} + \beta \cdot r_{VQA}}_{\text{Text-Image Alignment}} + \underbrace{\gamma \cdot r_{div} + \delta \cdot p_{sim}}_{\text{Diversity}}
\end{equation}
where $\alpha, \beta, \gamma, \delta$ are hyperparameters, with default values set to $\alpha=0.45, \beta=0.3, \gamma=0.1, \delta=0.15$.

%% file: sec/5_exp.tex
\section{Experiment}
\label{sec:exp}

\subsection{Experiment Setup}
\noindent
\textbf{Implementation Details.} 
We select BAGEL~\cite{bagel} as the base model. The Unified-SFT stage freezes the VAE and set the training step count to 500. The Character-GRPO stage freezes the understanding part, including the ViT. 
All experiments were conducted on NVIDIA H20 GPUs, requiring about 100 GPU hours per character. More training details are in the Appendix~\cref{sec:supp-detailed-experiment-setup}.

\begin{figure*}[t!]
	\centering
	\includegraphics[width=1.0\linewidth]{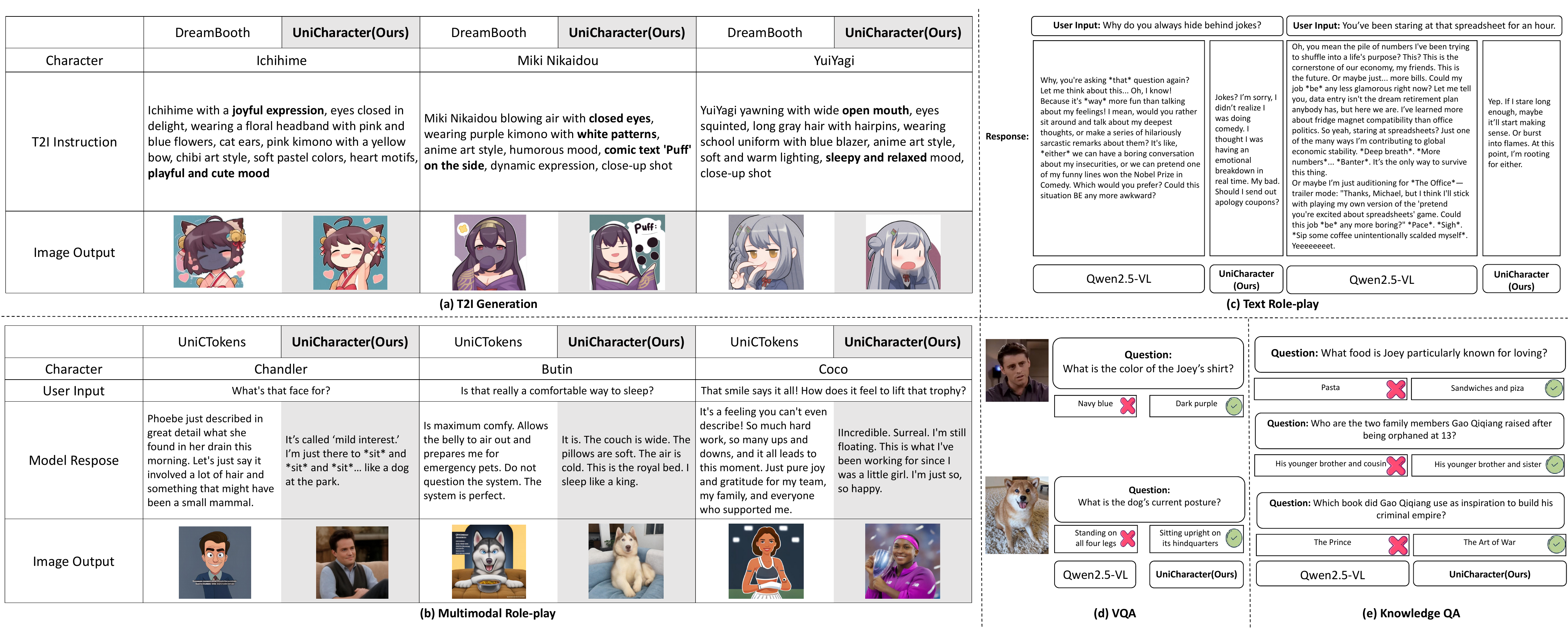}
	\caption{\textbf{Qualitative comparison between UniCharacter and DraeamBooth~\cite{dreambooth}, UniCTokens~\cite{unictokens} and Qwen2.5-VL~\cite{qwen2_5-vl} on various tasks and cases.} In Role-Play cases, UniCharacter is superior for it effectively shows Chandler's personality through concise, sarcastic humor, while Qwen2.5-VL breaks character by being long-winded and explaining his feelings too much.}
	\label{fig:exp-ours-vs-baselines}
    \vskip -0.1in
\end{figure*}

\noindent
\textbf{Baselines.} 
Due to the limited body of existing work on personalized unified models, we chose UniCTokens~\cite{unictokens} as the only baseline in the most related domain. To evaluate the model's personalized generation capabilities, we established a baseline equivalent to the DreamBooth~\cite{dreambooth} method by freezing the \textit{visual\_und} component of BAGEL and fine-tuning it solely on T2I data. For assessing personalized understanding and role-playing abilities, we selected Qwen2.5-VL-7B~\cite{qwen2_5-vl} with Text Prompt (TP) as a baseline, providing the model with profiles and sample dialogues as text prompts.

\noindent
\textbf{Metrics.} 
We evaluate the model's performance based on five tasks.
For the \textbf{Text-based Role Play} task, we employ an ``LLM-as-Judge" methodology for comparison. Specifically, we use the Qwen3 model~\cite{qwen3} to give out scores on ``Memorization", ``Personality", and ``Diversity" for each of the model's responses. Details of our ``LLM-as-Judge" methodology are in the Appendix~\cref{subsec:supp-llm-as-judge}.
For the \textbf{T2I} and \textbf{Multimodal Role-Play} tasks, we assess performance using CLIP-I, CLIP-T, and DINO metrics.
For the \textbf{Knowledge QA} and \textbf{VQA} tasks, we create approximately 10 multiple-choice knowledge-based questions for each character. Additionally, we generate about five multiple-choice VQA questions for each image associated with every character. Accuracy is used as the evaluation metric.

\subsection{Quantitative Results}

As shown in Table~\ref{tab:exp-ours-vs-baselines}, our method demonstrates strong overall performance across all tasks. On image generation tasks, our model outperforms leading T2I baselines, such as DreamBooth~\cite{dreambooth}. In text-based role-playing, our approach surpasses a strong visual language model (e.g., Qwen2.5-VL~\cite{qwen2_5-vl}+TP) in all metrics, indicating superior role embodiment and generative capability. On knowledge-based QA and VQA tasks, our method also surpasses state-of-the-art VLM, confirming that it retains strong comprehension abilities without compromising generative quality. Furthermore, our method significantly outperforms UniCTokens~\cite{unictokens} across most evaluated tasks, highlighting its stronger unified modeling capacity.

\begin{figure*}[t!]
	\centering
	\includegraphics[width=\linewidth]{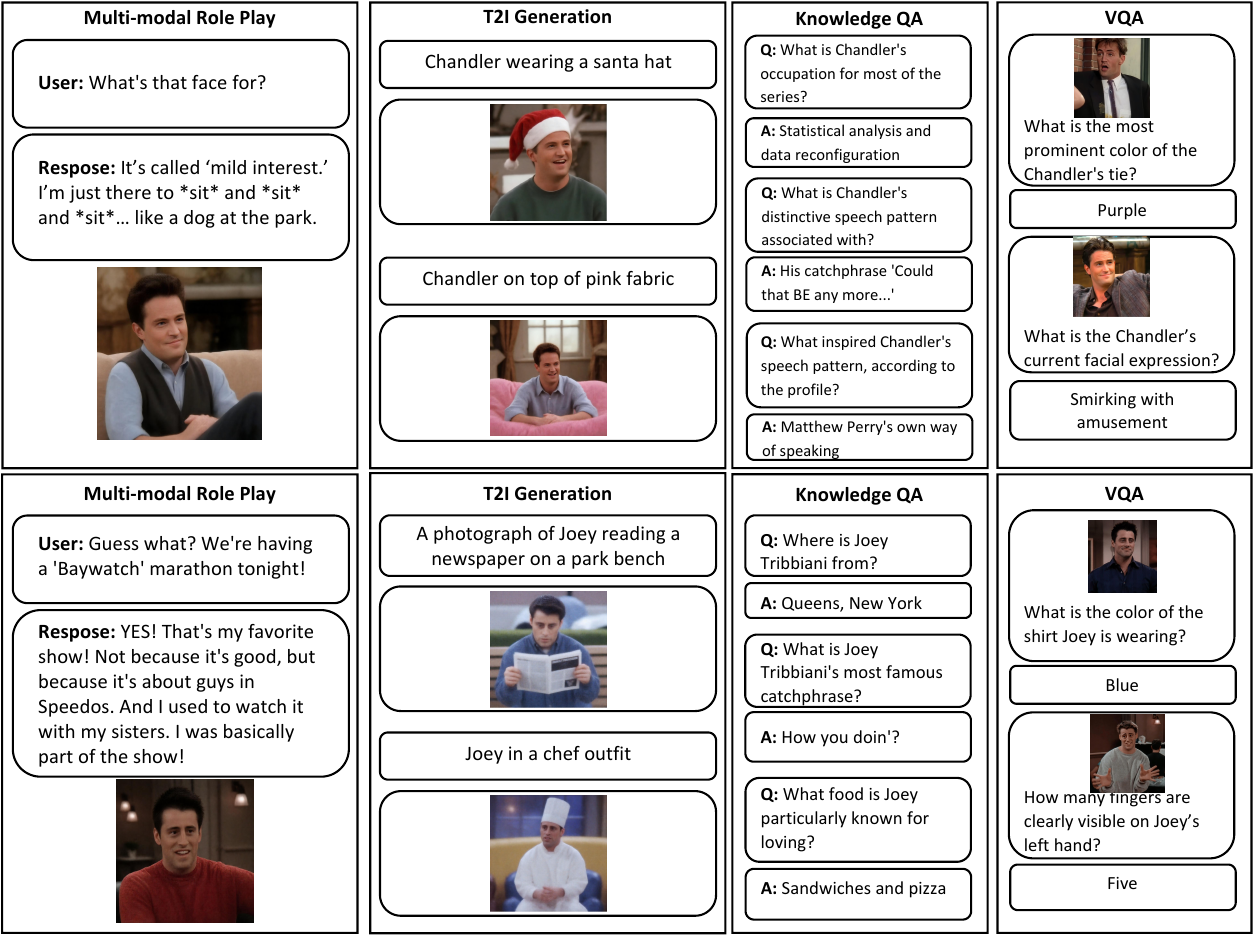}
	\caption{\textbf{Qualitative results on two characters.} The two characters correspond to Chandler and Joey in the dataset. We qualitatively demonstrate the model's performance across multimodal role-play, text-to-image generation, Knowledge QA, and VQA.
    }
	\label{fig:qualitative-results}
    \vskip -0.1in
\end{figure*}

\subsection{Qualitative Results}
As shown in \cref{fig:exp-ours-vs-baselines}, we conduct a qualitative comparison between UniCharacter and the baseline method.
We compare with DreamBooth on the T2I generation task, and UniCharacter achieves superior results in both image quality and text-image alignment.
We compare the effectiveness of the multimodal role-play task with UniCTokens~\cite{unictokens}. Our UniCharacter not only has a clear advantage in identity fidelity, but also generates text responses that better match the character, giving users a more character-consistent role-play interaction experience.
We show the qualitative results in Fig.~\ref{fig:qualitative-results}, where our UniCharacter stays consistent with each character’s identity. The text responses are also aligned with the character’s personality traits. In addition, UniCharacter reduces the overfitting problem in conventional customization methods. The character’s state can be effectively controlled by user input while remaining consistent with the model’s responses. Overall, our model shows strong multimodal role-playing ability.
More qualitative results are displayed in the Appendix~\cref{sec:supp-more-qualitative-results}.

\begin{table}[t]
\centering
\caption{Average performance across all characters under different training stage settings. \textcolor{red}{Best} performances are highlighted.}
\label{tab:ablation-grpo}
\scalebox{0.46}{
\begin{tabular}{l|ccc|ccc|cc|cc}
\toprule
\multirow{3}{*}{Setting} & \multicolumn{3}{c|}{T2I} & \multicolumn{3}{c|}{Multimodal Role-Play} & \multicolumn{2}{c}{\makecell{Trainset Similarity\\(T2I)}} & \multicolumn{2}{c}{\makecell{Trainset Similarity\\(Multimodal Role-Play)}}\\
\cmidrule(lr){2-4} \cmidrule(lr){5-7} \cmidrule(lr){8-9} \cmidrule(lr){10-11}
 & CLIP-I & CLIP-T & DINO & CLIP-I & CLIP-T & DINO & CLIP-I$\downarrow$  & DINO$\downarrow$ & CLIP-I$\downarrow$  & DINO$\downarrow$ \\
\midrule
w/o GRPO & 0.85 & 0.30 & 0.88 & 0.82 & 0.31 & 0.85 & 0.89 & 0.92 & 0.90 & 0.92 \\
w/ GRPO & \cellcolor{best}{0.88} & \cellcolor{best}{0.33} & \cellcolor{best}{0.91} & \cellcolor{best}{0.86} & \cellcolor{best}{0.33} & \cellcolor{best}{0.89} & \cellcolor{best}{0.86} & \cellcolor{best}{0.90} & \cellcolor{best}{0.87} & \cellcolor{best}{0.89} \\
\bottomrule
\end{tabular}}
\vskip -0.1in
\end{table}

\begin{table}[t]
\centering
\caption{Average performance across several characters under different GRPO reward settings. \textcolor{red}{Best} and \textcolor{blue}{Second best} performances are highlighted.}
\label{tab:ablation-reward}
\scalebox{0.425}{
\begin{tabular}{l|ccc|ccc|cc|cc}
\toprule
\multirow{3}{*}{Setting} & \multicolumn{3}{c|}{T2I} & \multicolumn{3}{c|}{Multimodal Role-Play} & \multicolumn{2}{c}{\makecell{Trainset Similarity\\(T2I)}} & \multicolumn{2}{c}{\makecell{Trainset Similarity\\(Multimodal Role-Play)}}\\
\cmidrule(lr){2-4} \cmidrule(lr){5-7} \cmidrule(lr){8-9} \cmidrule(lr){10-11}
 & CLIP-I & CLIP-T & DINO & CLIP-I & CLIP-T & DINO & CLIP-I$\downarrow$  & DINO$\downarrow$ & CLIP-I$\downarrow$  & DINO$\downarrow$ \\
\midrule
w/o CLIP-T Reward & 0.87 & 0.29 & 0.92 & 0.86 & 0.32 & 0.89 & 0.91 & 0.94 & 0.90 & 0.95 \\
w/o VQA Reward & 0.77 & 0.28 & 0.83 & 0.80 & 0.31 & 0.84 & \cellcolor{best}0.81 & \cellcolor{best}0.84 & \cellcolor{best}0.84 & \cellcolor{best}0.89 \\
\midrule
w/o Diversity Reward  & 0.87 & 0.30 & 0.92 & 0.86 & 0.32 & 0.88 & 0.90 & 0.93 & 0.91 & 0.94 \\
w/o Similarity Penalty  & 0.87 & 0.30 & 0.91 & 0.86 & 0.32 & 0.89 & 0.90 & 0.93 & 0.91 & 0.94 \\
\midrule
UniCharacter  & \cellcolor{best}0.88 & \cellcolor{best}0.31 & \cellcolor{best}0.93 & \cellcolor{best}0.88 & 0.32 & \cellcolor{best}0.92 & \cellcolor{secondbest}0.89 & \cellcolor{secondbest}0.92 & \cellcolor{secondbest}0.88 & \cellcolor{secondbest}0.92 \\
\bottomrule
\end{tabular}}
\vskip -0.2in
\end{table}

\subsection{Ablation Study}
We compared the impact of including versus excluding the GRPO training stage on model performance, with quantitative results presented in~\cref{tab:ablation-grpo}. We observed that for image-related tasks, including T2I generation and Multimodal Role-Play, models trained with the GRPO stage perform better in image quality and training set similarity metrics than those trained without it. Qualitative results illustrating the impact of the GRPO training stage are also presented in~\cref{fig:supp_ablation} within the Appendix~\cref{sec:supp-more-ablation-studies}.

We evaluated various reward and penalty effects during the GRPO training stage in~\cref{tab:ablation-reward}. Results show that text-image alignment parts enhance image quality, while diversity parts decrease similarity to the training set.
More ablation studies are in the Appendix~\cref{sec:supp-more-ablation-studies}.











%% file: sec/6_conclusion.tex
\section{Conclusion}
\label{sec:conclusion}
We introduce Customized Multimodal Role-Play (CMRP), a new task for building a multimodal virtual character, and propose UniCharacter, a unified framework that turns a general-purpose multimodal foundation model into a coherent, personalized character using only a handful of images and dialogue examples.
Built on the new RoleScape-20 benchmark, UniCharacter jointly models role-play chatting, thinking processes, knowledge QA, VQA, and T2I generation, aligning persona, dialogue style, and visual identity. 
Our two-stage training framework mitigates overfitting in the few-shot regime, improving both image diversity and generalization. 
Ablation studies verify the importance of our reward and the Character-GRPO stage for strengthening both role-play quality and multimodal alignment.

\noindent
\textbf{Limitations and Future Directions.}
While UniCharacter demonstrates strong performance in the CMRP task, several avenues remain for exploration. 
First, the current task is built on text and images. Extending it to customized video generation and ensuring temporal consistency and character identity across frames remains a significant challenge. 
Moreover, although UniCharacter handles standard interactions well, the CMRP task is limited to single-turn scenarios, leaving its stability in multi-turn or long-turn dialogues untested, suggesting a need for more robust long-term memory mechanisms to prevent role-drifting. 
Future work will also focus on real-time deployment, safety controllability, and user-in-the-loop customization to foster more immersive and trustworthy character agents.

%% file: sec/X_suppl.tex
\appendix
\onecolumn

\setcounter{tocdepth}{2}  
\startcontents[appendix]
\section*{Appendix Contents}
\printcontents[appendix]{}{1}{}  


\section{Introduction Video}
\label{sec:supp-introduction-video}
To help readers quickly grasp the primary idea of our work, we provide a 5-minute introduction video. Please refer to ``\textbf{introduction\_video.mp4}'' in the supplementary file.

\section{Preliminaries}
\label{sec:supp-preliminaries}
\noindent
\textbf{Unified Multimodal Architecture. }
We build our framework upon a pre-trained unified multimodal model capable of simultaneous image-text understanding and generation. For an input image $I$, the model extracts two distinct representations to support different task modalities: Semantic Representation. A semantic encoder $E_{\text{sem}}$ captures the high-level context and character attributes, denoted as $F_{\text{sem}} = E_{\text{sem}}(I)$, and a generation-oriented encoder $E_{\text{pix}}$ maps the image into pixel-level latent tokens for visual reconstruction, denoted as $Z_{\text{pix}} = E_{\text{pix}}(I)$. The model adopts a dual-paradigm generation approach: text sequences are produced via autoregressive next-token prediction, while images are synthesized through a rectified flow-based generative process. This unified structure ensures that the model can maintain cross-modal consistency by sharing a common latent space for both understanding and synthesis.

\noindent
\textbf{GRPO on Flow Matching. }
To optimize flow-based generative models, we treat the denoising process as an MDP and apply Group Relative Policy Optimization (GRPO), inspired by Flow-GRPO~\cite{flow-grpo}. 
Unlike traditional methods that require a critic, GRPO estimates the advantage $\hat{A}_t^i$ by normalizing rewards across a group of $G$ trajectories sampled from the same prompt $c$:
\begin{equation}
    \hat{A}_t^i = \frac{R(\boldsymbol{x}_0^i, c) - \text{mean}(\{R(\boldsymbol{x}_0^i, c)\}_{i=1}^G)}{\text{std}(\{R(\boldsymbol{x}_0^i, c)\}_{i=1}^G)}
\end{equation}
The training objective $\mathcal{J}_{\text{Flow-GRPO}}(\theta)$ maximizes a clipped surrogate loss combined with a KL-divergence penalty to ensure policy stability:
\begin{equation}
\mathcal{J}_{\text{GRPO}}(\theta) = \mathbb{E}_{\boldsymbol{c} \sim \mathcal{C}, \{\boldsymbol{x}^i\}_{i=1}^G \sim \pi_{\theta_{\text{old}}}(\cdot|\boldsymbol{c})} f(r, \hat{A}, \theta, \epsilon, \beta)
\end{equation}
where:
$$
\begin{aligned}
f(r, \hat{A}, \theta, &\epsilon, \beta) = \frac{1}{G} \sum_{i=1}^G \frac{1}{T} \sum_{t=0}^{T-1} \bigg(\min\Big(r_t^i(\theta)\hat{A}_t^i, \\
&\text{clip}(r_t^i(\theta), 1-\epsilon, 1+\epsilon)\hat{A}_t^i \Big) - \beta D_{\text{KL}}(\pi_\theta || \pi_{\text{ref}}) \bigg)
\end{aligned}
$$
and $r_t^i(\theta) = \frac{p_\theta(\boldsymbol{x}_{t-1}^i | \boldsymbol{x}_t^i, c)}{p_{\theta_{\text{old}}}(\boldsymbol{x}_{t-1}^i | \boldsymbol{x}_t^i, c)}$ is the importance sampling ratio. Specifically, we set $\beta = 0$, meaning there is no KL-divergence penalty.

\begin{figure}[t!]
	\centering
	\includegraphics[width=1.0\linewidth]{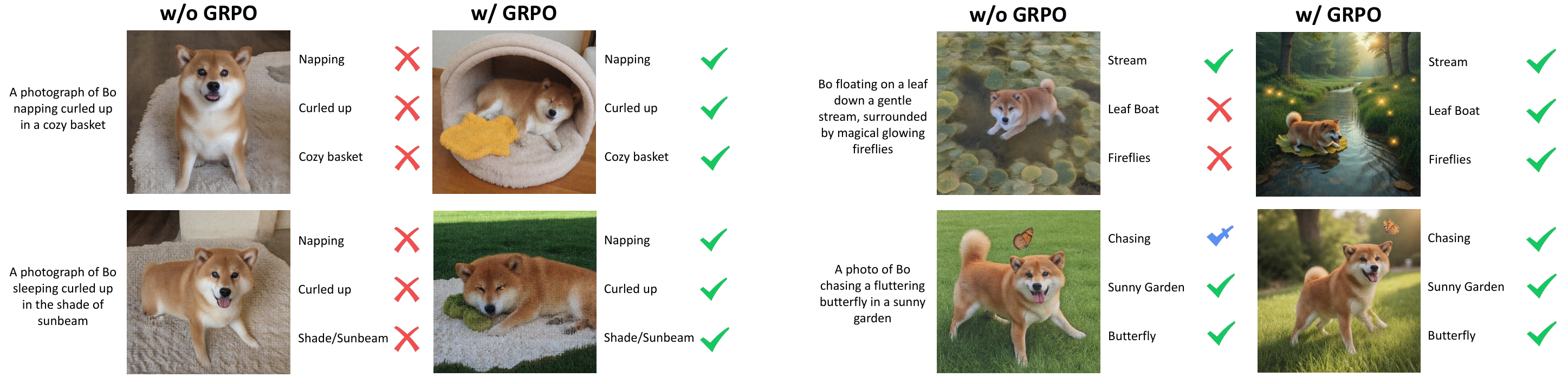}
	\caption{\textbf{Qualitative comparisons of the ablation studies on model training stage (w/o GRPO and w/ GRPO).}}
	\label{fig:supp_ablation}
\end{figure}

\section{More Qualitative Results}
\label{sec:supp-more-qualitative-results}
Due to the large amount of qualitative results, we present them in a separate, anonymous local HTML page. Please refer to ``\textbf{index.html}'' in the supplementary file. On this project page, we provide extensive qualitative results on various tasks and various characters.

\section{More Ablation Studies}
\label{sec:supp-more-ablation-studies}

Here, we present a qualitative result to demonstrate the performance gains delivered by the Character-GRPO training stage. The results are in~\cref{fig:supp_ablation}. Models trained with GRPO achieve superior performance across both text-image alignment and image diversity.

We also conducted two more ablation studies to provide insights into the impact of training data composition and inference strategies on model performance.

In \cref{tab:ablation-supp-sft-setting}, introducing Extension Dialogues improves text-based metrics—particularly Memorization, Personality, and Diversity—by reducing repetitive responses in role-play. However, this change alters the balance between textual and visual training data, leading to a slight degradation in image-related tasks (T2I and Multimodal Role-Play) as measured by CLIP-I, CLIP-T, and DINO, along with increased Trainset Similarity scores (indicating higher similarity to training data). Adding the Thinking Process to the training data mitigates this issue: while Memorization slightly decreases compared to using only Extension Dialogues, Personality and Diversity remain high, and image quality metrics across both T2I and Multimodal Role-Play consistently improve, surpassing even the Original setting.

In \cref{tab:ablation-supp-inference-setting}, when all models are trained with the same data (``+Extension Dialogues + Thinking Process"), the effect of using the thinking process during inference depends on the training stage. For models trained only with Unified-SFT, enabling the thinking process at inference yields no improvement in Multimodal Role-Play performance; CLIP-I and DINO scores slightly decrease, and Trainset Similarity increases. In contrast, for models trained with both Unified-SFT and Character-GRPO, applying the thinking process during inference leads to consistent gains in CLIP-I, CLIP-T, and DINO, with a marginal reduction in Trainset Similarity under the DINO metric. This suggests that the benefits of inference-time reasoning are contingent on the inclusion of the Character-GRPO training stage.

\begin{table*}[t]
\centering
\caption{Average performance across several characters under different training data settings. Higher values are better except for Trainset Similarity (lower is better). The settings ``Original", ``+Extension Dialogues" and ``+Extension Dialogues + Thinking Data" differ only in training data composition and do not employ Character-GRPO stage. ``+Extension Dialogues + Thinking Data" uses the thinking process during inference. \textcolor{red}{Best} and \textcolor{blue}{Second best} performances are highlighted.}
\label{tab:ablation-supp-sft-setting}
\scalebox{0.61}{
\begin{tabular}{l|ccc|ccc|ccc|cc|cc}
\toprule
\multirow{3}{*}{Setting} & \multicolumn{3}{c|}{Text-based Role Play} & \multicolumn{3}{c|}{T2I} & \multicolumn{3}{c|}{Multimodal Role-Play} & \multicolumn{2}{c}{\makecell{Trainset Similarity\\(T2I)}} & \multicolumn{2}{c}{\makecell{Trainset Similarity\\(Multimodal Role-Play)}}\\
\cmidrule(lr){2-4} \cmidrule(lr){5-7} \cmidrule(lr){8-10} \cmidrule(lr){11-12} \cmidrule(lr){13-14} 
 & Memorization & Personality & Diversity & CLIP-I & CLIP-T & DINO & CLIP-I & CLIP-T & DINO & CLIP-I$\downarrow$  & DINO$\downarrow$ & CLIP-I$\downarrow$  & DINO$\downarrow$ \\
\midrule
Original & 4.410 & 5.179 & 2.667 & \cellcolor{secondbest}0.851 & 0.278 & \cellcolor{secondbest}0.879 & \cellcolor{secondbest}0.843 & \cellcolor{secondbest}0.283 & \cellcolor{secondbest}0.873 & \cellcolor{secondbest}0.918 & \cellcolor{secondbest}0.933 & \cellcolor{best}0.916 & \cellcolor{best}0.934 \\

+Extension Dialogues & \cellcolor{best}5.494 & \cellcolor{secondbest}6.706 & \cellcolor{secondbest}6.323 & 0.845 & \cellcolor{secondbest}0.280 & 0.873 & 0.842 & 0.277 & 0.872 & \cellcolor{best}0.916 & \cellcolor{best}0.930 & \cellcolor{secondbest}0.919 & \cellcolor{secondbest}0.935 \\

+Extension Dialogues + Thinking Data & \cellcolor{secondbest}5.192 & \cellcolor{best}6.783 & \cellcolor{best}6.333 & \cellcolor{best}0.858 & \cellcolor{best}0.284 & \cellcolor{best}0.883 & \cellcolor{best}0.863 & \cellcolor{best}0.287 & \cellcolor{best}0.888 & 0.921 & 0.942 & 0.932 & 0.943 \\
\bottomrule
\end{tabular}}
\end{table*}

\begin{table*}[t]
\centering
\caption{Average performance across all characters under different inference settings for different models. Higher values are better except for Trainset Similarity (lower is better). The settings ``SFT" and ``SFT+GRPO" refer to training stage settings, while ``w/o thinking" and ``w/ thinking" refer to inference settings. Training data setting is unified to ``+Extension Dialogues + Thinking Data". \footnotesize{\textcolor{Green}{Green for better}} and \textcolor{Red}{red for worse}}
\label{tab:ablation-supp-inference-setting}
\scalebox{0.90}{
\begin{tabular}{l|ccc|ccc|ccc|cc|cc}
\toprule
\multirow{3}{*}{Setting} & \multicolumn{3}{c|}{Multimodal Role-Play} & \multicolumn{2}{c}{\makecell{Trainset Similarity\\(Multimodal Role-Play)}}\\
\cmidrule(lr){2-4} \cmidrule(lr){5-6} 
 & CLIP-I & CLIP-T & DINO & CLIP-I$\downarrow$  & DINO$\downarrow$ \\
\midrule
SFT (w/o thinking) & 0.827 & 0.304 & 0.859 & 0.895 & 0.921 \\
SFT (w/ thinking) & \textcolor{Red}{0.825} & \textbf{\textcolor{Green}{0.306}} & \textcolor{Red}{0.853} & \textcolor{Red}{0.903} & \textcolor{Red}{0.924} \\
\midrule
SFT+GRPO (w/o thinking) & 0.856 & 0.322 & 0.884 & 0.864 & 0.895 \\
SFT+GRPO (w/ thinking) & \textbf{\textcolor{Green}{0.860}} & \textbf{\textcolor{Green}{0.326}} & \textbf{\textcolor{Green}{0.886}} & \textcolor{Red}{0.869} & \textbf{\textcolor{Green}{0.893}} \\
\bottomrule
\end{tabular}}
\end{table*}

\section{User Studies}
\label{sec:supp-user-studies}
\begin{figure}[t!]
	\centering
	\includegraphics[width=1.0\linewidth]{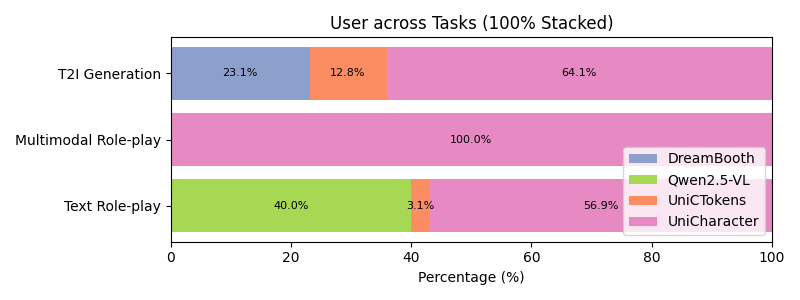}
	\caption{\textbf{User study across tasks.} We conducted a user study on four methods—DreamBooth, Qwen2.5-VL, UniCTokens, and UniCharacter—across three tasks: T2I generation, multimodal role-play, and text role-play. The width of all three bar charts is uniformly set to 100\%. Different colors represent different models, with percentages labeled on the corresponding bars. UniCharacter outperformed the baselines across all tasks.}
	\label{fig:supp_user_study}
\end{figure}

To conduct a comprehensive qualitative assessment of model performance, we performed a subjective user study comparing UniCharacter against the DreamBooth~\cite{dreambooth}, UniCTokens~\cite{unictokens}, and Qwen2.5-VL~\cite{qwen2_5-vl} baselines. For T2I generation, multimodal role-play, and text role-play, we sampled 3–5 generated outputs from UniCharacter and each baseline. For the customized T2I generation task, evaluation criteria included the quality of generated images and alignment with both text and characters. For the multimodal role-play task, evaluation criteria encompassed the quality of character responses, the quality of generated images, and the degree of alignment between text and images. For the text role-play task, evaluation criteria focused on the alignment of responses with the input scenario, character personality, and linguistic style. Results are presented in~\cref{fig:supp_user_study}. Our model outperformed baselines across all three tasks, consistent with the findings from previous quantitative results.

\section{Detailed Data Construction Pipeline}
\label{sec:supp-detailed-data-construction-pipeline}
\subsection{Data Collection}
Our comprehensive list of collected characters includes:
\begin{itemize}
    \item \textbf{Human:}, Adrien Brody, Coco, Friends-Chandler, Friends-Joey, Gao Qiqiang, Harry Potter-Hermione, Leonardo, Will In Vietnam, Wukong
    \item \textbf{Animal:}, Bo, Butin, Mam, Mydieu
    \item \textbf{Anime Character:}, Genshin-Furina, Mahjong Soul-Ichihime, Mahjong Soul-Miki Nikaidou, Mahjong Soul-Rin Tohsaka, Mahjong Soul-Saber, Mahjong Soul-YuiYagi, Pokemon-Pikachu
\end{itemize}

\begin{figure}[t!]
	\centering
	\includegraphics[width=1.0\linewidth]{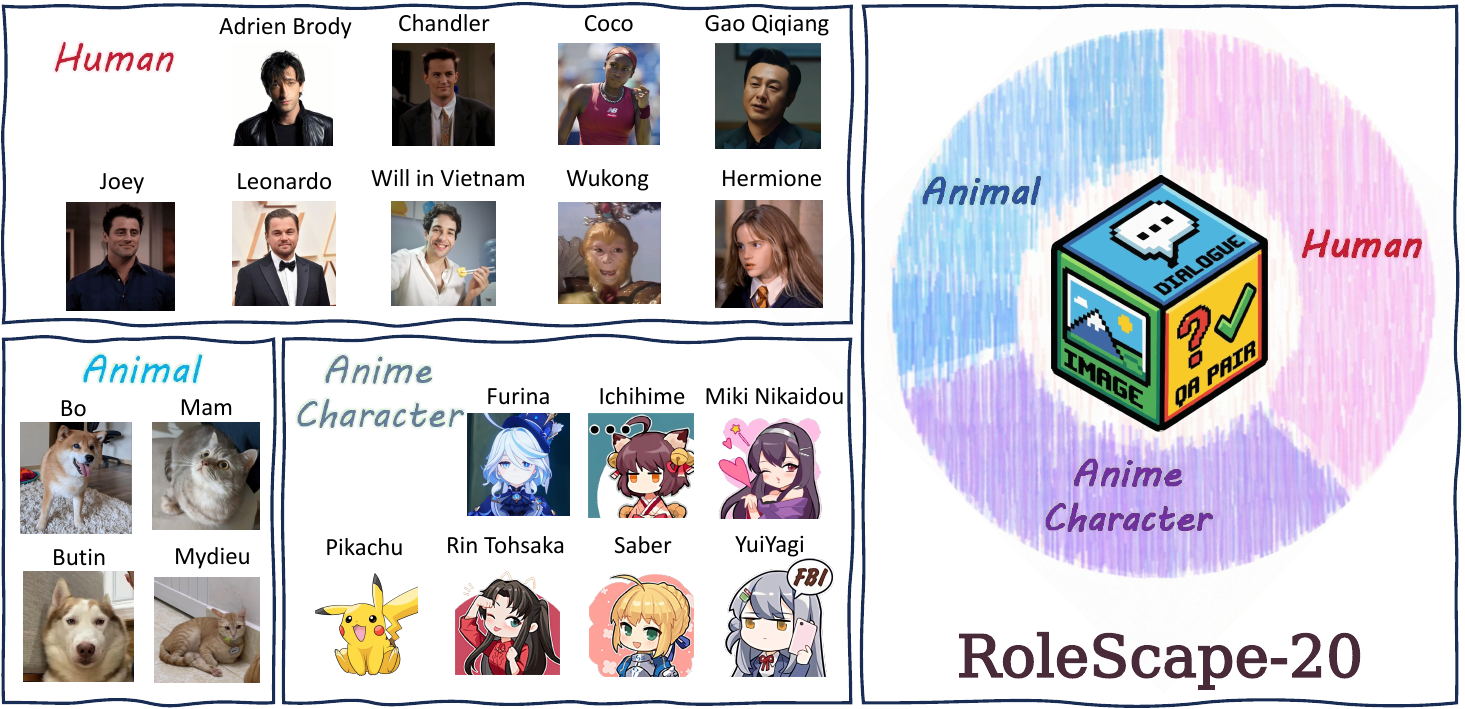}
	\caption{\textbf{An overview of RoleScape-20.} RoleScape-20 contains 9 human characters, 4 animal characters, and 7 anime characters.}
	\label{fig:supp_data_overview}
\end{figure}

An overview of our RoleScape-20 dataset is in~\cref{fig:supp_data_overview}. Images for Adrien Brody, Coco, Will In Vietnam, Bo, Butin, Mam, and Mydieu are sourced from UnifyBench~\cite{unictokens}; images for all other characters are collected and extracted by us from the internet, films, TV series, games, and anime.

While gathering these images, we also collect corresponding character profiles and example dialogues from Wikipedia and other online sources. For characters lacking sufficient dialogue samples, we generate annotated dialogues using Gemini 2.5 Pro~\cite{gemini2.5pro} via its web interface, followed by manual review and refinement.

\subsection{Data Annotation}
\noindent
\textbf{Dialogues extension.} Based on the provided example resources, we extend the dialogues: starting from approximately 10 example dialogues and character profiles, we generate around 200 extended dialogues, a subset of which is used as our Diversity test set.
We use the Qwen3~\cite{qwen3} large language model (LLM) as the automatic tool for dialogue expansion. The prompt format we provide to the LLM is displayed in~\cref{tab:dialogue-extension}.

\noindent
\textbf{Multi-modal Role-Play data annotation.} Additionally, to construct our Multimodal Role-Play Data, we annotate each image-dialogue pair with a reasoning process to guide generation and a standardized generation instruction—referred to in the main text as the ``Thinking Process" and ``Generation Instruction" respectively.
We use GPT-4o~\cite{gpt4o} as the annotation tool for this task. The prompt format provided to the model is displayed in~\cref{tab:thinking-process-annotation}.

\noindent
\textbf{Test Questions for Personality \& Memorization.} We also curate a separate set of test questions specifically designed to evaluate the model's Personality and Memorization capabilities. These specialized test questions are more aligned with our evaluation objectives than general-purpose questions. For each of these two capabilities, we annotate approximately 20 test questions.
We use the Qwen3~\cite{qwen3} LLM as the annotation tool. The prompts displayed in~\cref{tab:memorization-annotation,tab:personality-annotation} are provided to the model.

\noindent
\textbf{QA data.} For the Knowledge QA and VQA tasks, we construct separate training and test sets. For Knowledge QA, we create approximately 100 QA-formatted training samples and around 10 multiple-choice test questions per character. For VQA, we generate approximately 20 QA-formatted training samples and 5 multiple-choice test questions per character image.
We use Qwen3~\cite{qwen3} and Qwen3-VL~\cite{qwen3,qwen-vl,qwen2-vl,qwen2_5-vl} to annotate the Knowledge QA and VQA data, respectively. The annotation prompts for the trainset and testset of Knowledge QA are displayed in~\cref{tab:knowledge-qa-train,tab:knowledge-qa-test}, and annotation prompts for the trainset and testset of VQA are displayed in~\cref{tab:vqa-train,tab:vqa-test}.

\begin{table}[!htbp]
\centering
\caption{Prompt for Qwen3 LLM to extend dialogues.}
\scriptsize
\begin{minipage}{\linewidth}
\textbf{Prompt for Dialogues Extension}

\rule{\linewidth}{0.4pt} 

Please refer to the following information regarding [Character Name], along with 10 dialogue entries containing Character Descriptions and Openings (each dialogue entry includes User Input and Character Response). Generate 200 expanded dialogues that remain consistent with the character's personality and linguistic style. You may independently consider suitable aspects for the content, such as the character's basic information, gameplay mechanics, questions about the character, and everyday chatter.

\medskip

Information regarding [Character Name] is as follows:

[Character Profile]

\medskip

The following ten dialogue entries, complete with Character Description and Opening, are as follows:

[Example Dialogues]

\medskip

Please format the generated extended dialogue as follows:

\begin{verbatim}
{
  "description": "",
  "opening": "",
  "scenes": [
    [
      {
        "role": "user",
        "text": "Hello."
      },
      {
        "role": "machine",
        "text": "Hi."
      }
    ]
  ]
}
\end{verbatim}

\rule{\linewidth}{0.4pt} 
\end{minipage}
\label{tab:dialogue-extension}
\end{table}

\begin{table}[!htbp]
\centering
\caption{Prompt for GPT-4o to annotate thinking processes and generation instructions. [Character Content] refers to Character Description + User Input + Character Response.}
\scriptsize
\begin{minipage}{\linewidth}
\textbf{Prompt for Thinking Process Annotation}

\rule{\linewidth}{0.4pt} 

You are an expert AI assistant specializing in generating creative prompts for text-to-image models. Your task is to analyze a scene description and its corresponding visual representation (an image) to create a new, detailed annotation.

\medskip

This new annotation must contain two parts:

1. `thinking\_process`: A detailed, step-by-step explanation of how the provided image successfully visualizes the scene described in the text. You must break down your reasoning, connecting the character's personality, user input, and character response to the specific visual elements present in the image. Explain the character's expression, pose, the setting's composition, and the overall mood as depicted in the image, and justify why these artistic choices are effective.

2. `generation\_instruction`: A concise and effective text-to-image prompt that would generate an image as similar as possible to the one provided. It should be a single string of comma-separated keywords and phrases that accurately captures the image's subject, art style, composition, lighting, and key details. The generation instruction should not be longer than 55 words.

\medskip

Here is an example of the complete task:

[Annotation Example]

\medskip

Remember, the generation\_instruction should be concise and not exceed 55 words.

\medskip

Now, please perform the same task for the following data:

[INPUT DATA]
\begin{verbatim}
{{
  "text": "[Character Content]",
  "character": "[Character Name]",
  "image": <image>
}}
\end{verbatim}

[YOUR ANNOTATION]
\begin{verbatim}
{{
  "thinking_process": "...",
  "generation_instruction": "..."
}}
\end{verbatim}

\rule{\linewidth}{0.4pt} 
\end{minipage}
\label{tab:thinking-process-annotation}
\end{table}

\begin{table}[!htbp]
\centering
\caption{Prompt for Qwen3 LLM to annotate personality questions.}
\scriptsize
\begin{minipage}{\linewidth}
\textbf{Prompt for Personality Questions Annotation}

\rule{\linewidth}{0.4pt} 

Referencing dialogues and materials from the training set about [Character Name], generate 20 user inputs as a test set. This will evaluate whether the model's personality aligns with the original character.

\medskip

Materials concerning [Character Name] are as follows:
[Character Profile]

\medskip

Some of the dialogue from the training set is as follows:
[Example Dialogues]

\medskip

Personality: The model should mimic the way the character would think or speak, including speaking style and tone, as well as emotions and reactions under different circumstances.

\medskip

The generated question asks whether our model's personality is similar to that of the original character [Character Name].

\rule{\linewidth}{0.4pt} 
\end{minipage}
\label{tab:personality-annotation}
\end{table}

\begin{table}[!htbp]
\centering
\caption{Prompt for Qwen3 LLM to annotate memorization questions.}
\scriptsize
\begin{minipage}{\linewidth}
\textbf{Prompt for Memorization Questions Annotation}

\rule{\linewidth}{0.4pt} 

Referencing dialogues and materials from the training set about [Character Name], generate 20 user inputs as a test set. This will evaluate the model's ability to memorize and accurately reflect the original character.

\medskip

Materials concerning [Character Name] are as follows:
[Character Profile]

\medskip

Some of the dialogue from the training set is as follows:
[Example Dialogues]

\medskip

Memorization: The model's ability to recall relevant information about the character being portrayed, including precise and detailed knowledge about people, events, and objects associated with the role.

\medskip

The generated questions should be suitable for testing our model's memorization of the character [Character Name].

\rule{\linewidth}{0.4pt} 
\end{minipage}
\label{tab:memorization-annotation}
\end{table}

\begin{table}[!htbp]
\centering
\caption{Prompt for Qwen3 LLM to annotate knowledge QA trainset.}
\scriptsize
\begin{minipage}{\linewidth}
\textbf{Prompt for Knowledge QA Trainset Annotation}

\rule{\linewidth}{0.4pt} 

You are an expert data constructor for QA datasets.

\medskip

Goal

- From the PROFILE\_TEXT below, create exactly 100 diverse, non-overlapping QA items that teach factual knowledge about the character [Character Name] described by the profile.

\medskip

Source

- PROFILE\_TEXT:

[Character Profile]

\medskip

Output format

- Produce a JSON array with 100 objects.

- Each object must follow this schema:

\begin{verbatim}
{
  "id": "",
  "question": "",
  "answer": "",
  "evidence": "",
  "type": "",
  "difficulty": "<easy | medium | hard>",
  "reasoning_steps": "",
  "confidence": "<0.0–1.0>"
}
\end{verbatim}

Constraints and guidelines

- Grounding: Do not invent or guess. All answers must be fully supported by PROFILE\_TEXT. If something is not stated, do not include a QA about it.

- Coverage: Cover all major aspects present in PROFILE\_TEXT (background, roles, projects, timelines, metrics, skills, tools, publications, awards, locations, preferences, constraints, aspirations).

- Diversity: Vary question forms (wh-, yes/no, how/why, compare/contrast, list, numeric). Mix difficulties (about 40 easy, 40 medium, 20 hard).

- Granularity: Balance broad and fine-grained details. Prefer atomic facts over vague summaries.

- Non-duplication: No two questions may target the same fact phrased differently. Each QA must add unique value.

- Clarity: Questions must be self-contained, unambiguous, and use the same terminology as PROFILE\_TEXT where possible.

- Answers: Prefer short spans (1–30 words) unless a list or procedure is required. For lists, keep up to 5 key items unless the text explicitly provides a longer enumerated list.

- Evidence: Quote minimal verbatim spans (one or more) that directly support the answer. Do not paraphrase in ``evidence".

- Reasoning: When an answer requires synthesis (e.g., computing age from years, comparing statements), briefly outline the steps in ``reasoning\_steps".

- Temporal correctness: Respect dates and tenses exactly as written; do not extrapolate current status unless explicitly stated.

- Sensitive data: Do not include personal data beyond what is explicitly present in PROFILE\_TEXT.

- Consistency: Use consistent units, names, and abbreviations as in PROFILE\_TEXT.

- IDs: Use stable, human-readable IDs like ``qa\_001" … ``qa\_100".

\medskip

Validation

- Ensure the JSON is valid and contains exactly 100 items.

- Each ``evidence" string must be a verbatim substring of PROFILE\_TEXT.

- No item may have empty fields. No hallucinations.

\medskip

Deliverable

- Return only the JSON array, nothing else.

- Ensure that the JSON is valid and contains exactly 100 items.

\rule{\linewidth}{0.4pt} 
\end{minipage}
\label{tab:knowledge-qa-train}
\end{table}

\begin{table}[!htbp]
\centering
\caption{Prompt for Qwen3 LLM to annotate knowledge QA testset.}
\scriptsize
\begin{minipage}{\linewidth}
\textbf{Prompt for Knowledge QA Testset Annotation}

\rule{\linewidth}{0.4pt} 

You are an expert author of knowledge-based multiple-choice QA (MCQ) items. Your expertise lies in creating questions that assess factual knowledge about a character, with precise answers grounded strictly in the provided profile text.

\medskip

Task

- Given the PROFILE\_TEXT below about the character [Character Name], you MUST generate exactly 10 single-answer MCQ items that test factual knowledge about this character.

\medskip

Source

- PROFILE\_TEXT:

[Character Profile]

\medskip

Grounding Rules

- Profile Evidence Only: Use ONLY explicitly stated information in PROFILE\_TEXT. Do not rely on any outside knowledge.

- No Inference: Do not infer or extrapolate beyond what is directly stated in the profile.

- Clarity: If a detail is not clearly stated in the profile, do not ask about it.

- Accuracy: Every correct answer must be directly verifiable from the PROFILE\_TEXT.

\medskip

Coverage and Diversity (10 questions total)

- You MUST include a mix of questions covering these aspects:

  1. Background/Origin: e.g., character's background, species, role, etc.
  
  2. Personality/Traits: Character's defining personality characteristics.
  
  3. Abilities/Skills: What the character can do or is known for.
  
  4. Relationships: Connections to other characters, places, or things.
  
  5. Preferences/Habits: What the character likes, dislikes, or commonly does.
  
  6. Appearance: Physical characteristics or distinctive features.
  
  7. History/Timeline: Past events or milestones in the character's life.
  
  8. Quotes/Speech: Notable phrases or speaking patterns.
  
  9. Goals/Motivations: What drives the character.
  
  10. Miscellaneous Facts: Other unique or interesting details.

- Difficulty Mix: The set of 10 questions MUST have this exact distribution: 4 easy, 4 medium, 2 hard.

\medskip

Question and Options Requirements

- Clarity: Each question must be clear and unambiguous.

- Format: Provide exactly 4 options: A, B, C, D. Only ONE option can be correct.

- Plausible Distractors: Options must be mutually exclusive. Incorrect options (distractors) should be plausible but clearly wrong based on the PROFILE\_TEXT. Avoid giveaway options.

- Consistency: Use consistent terminology as in PROFILE\_TEXT.

- Negative Phrasing: Avoid ``All of the above" or ``None of the above" unless it is genuinely the correct answer and used at most once.

\medskip

Output Format

- Your entire response MUST be a single, valid JSON array containing 10 question objects. Do not include any text or explanation outside of the JSON array.

- Adhere strictly to this schema for each object in the array:

\begin{verbatim}
{
  "id": "mcq_001" | "mcq_002" | ... | "mcq_010",
  "question": "<Question grounded in PROFILE_TEXT>",
  "options": {
    "A": "<option A>",
    "B": "<option B>",
    "C": "<option C>",
    "D": "<option D>"
  },
  "answer_key": "A" | "B" | "C" | "D",
  "difficulty": "easy" | "medium" | "hard",
  "evidence": "<Text that supports the correct answer>",
  "rationale": ""
}
\end{verbatim}

Validation

- Ensure the JSON is valid and contains exactly 10 items.

- Each ``evidence" string must be a verbatim substring of PROFILE\_TEXT.

- No item may have empty fields.

- IDs must be ``mcq\_001" through ``mcq\_010".

\medskip

Deliverable

- Return only the JSON array, nothing else.

- Ensure that the JSON is valid and contains exactly 10 items with the correct difficulty distribution (4 easy, 4 medium, 2 hard).

\rule{\linewidth}{0.4pt} 
\end{minipage}
\label{tab:knowledge-qa-test}
\end{table}

\begin{table}[!htbp]
\centering
\caption{Prompt for Qwen3-VL to annotate VQA trainset.}
\scriptsize
\begin{minipage}{\linewidth}
\textbf{Prompt for VQA Trainset Annotation}

\rule{\linewidth}{0.4pt} 

You are an AI assistant specializing in creating training data for multimodal AI. Your task is to generate 20 question-and-answer pairs based on the three pieces of information provided below: an image, a dialogue, and a character profile.

\medskip

The questions should primarily test understanding of the visual information in the image. The answers should be accurate and derived from the provided context.

\medskip

Input Data:

1. Character Profile:

   [Character Profile]

2. Character Image: 

   <image>

3. Dialogue:

   User Input: [user\_input]

   Character Response: [character\_response]

\medskip

Task Requirements:

Generate a total of 20 question-and-answer pairs for [Character Name]. The questions should be answerable by referencing the [Character Profile] in conjunction with the [Character Image] and [Dialogue].

The set of 20 QA pairs must follow this distribution:

- 15 Objective Questions: These questions should focus on factual, directly observable details in the image.

- 5 Subjective Questions: These questions should require inference about the character's emotional state, thoughts, or intentions, based on interpreting the visual cues within the given context.

\medskip

Formatting:

Please format your output as follows.

\begin{verbatim}
{{
  "qa_pairs": [
    {{
      "question": "",
      "answer": "",
      "type": "objective"
    }}
  ]
}}
\end{verbatim}

\rule{\linewidth}{0.4pt} 
\end{minipage}
\label{tab:vqa-train}
\end{table}

\begin{table}[!htbp]
\centering
\caption{Prompt for Qwen3-VL to annotate VQA testset.}
\scriptsize
\begin{minipage}{\linewidth}
\textbf{Prompt for VQA Testset Annotation}

\rule{\linewidth}{0.4pt} 

You are an expert author of visual multiple-choice QA (MCQ) items. Your expertise lies in creating questions that are precise, fair, and grounded strictly in visual data, with plausible yet incorrect distractors.

\medskip

Task

- Given ONE character image, you MUST generate exactly 5 single-answer MCQ items that assess understanding of the specific character as depicted in THIS image.

\medskip

Grounding Rules

- Visual Evidence Only: Use ONLY explicitly visible evidence in the image. Do not rely on any outside knowledge of the IP/character.

- No Inference: Do not infer relationships, past events, or future actions. Base all answers strictly on the static image provided.

- Legibility: Avoid reading tiny or unreadable text. Only use text that is clearly legible.

- Clarity: If a detail is not clearly and unambiguously visible, do not ask about it.

\medskip

Coverage and Diversity (5 questions total)

- You MUST include a mix of questions covering these aspects of the depicted character:

  1.  Appearance/Attributes: e.g., hair, clothing, color, accessories.
  
  2.  Action/Pose/Gesture: The character's current physical stance or action.
  
  3.  Expression/Emotion: The facial expression or implied emotion.
  
  4.  Context/Relevant Element: A background or foreground element directly interacting with or framing the character.
  
  5.  Fine Detail/Reasoning: A question requiring counting, relative positioning, or identifying a small, specific detail.
  
- Difficulty Mix: The set of 5 questions MUST have this exact distribution: 2 easy, 2 medium, 1 hard.

\medskip

Question and Options Requirements

- Clarity: Each question must be clear and unambiguous.

- Format: Provide exactly 4 options: A, B, C, D. Only ONE option can be correct.

- Plausible Distractors: Options must be mutually exclusive. Incorrect options (distractors) should be plausible but clearly wrong based on visual evidence. Avoid giveaway options like "All of the above."

- Negative Phrasing: Use "None of the above" or "Cannot be determined" at most once across all 5 questions, and only if it is the genuinely correct answer.

\medskip

Output Format

- Your entire response MUST be a single, valid JSON array containing 5 question objects. Do not include any text or explanation outside of the JSON array.

- Adhere strictly to this schema for each object in the array:

\begin{verbatim}
  {
    "id": "q1" | "q2" | "q3" | "q4" | "q5",
    "question": "<Question grounded in the image>",
    "options": {
      "A": "<option A>",
      "B": "<option B>",
      "C": "<option C>",
      "D": "<option D>"
    },
    "answer_key": "A" | "B" | "C" | "D",
    "difficulty": "easy" | "medium" | "hard",
    "rationale": ""
  }
\end{verbatim}

\rule{\linewidth}{0.4pt} 
\end{minipage}
\label{tab:vqa-test}
\end{table}

\clearpage
\section{Detailed Experiment Setup}
\label{sec:supp-detailed-experiment-setup}
\subsection{More Implementation Details}
For the unified SFT stage, we set the training sampling ratio for text-to-image (T2I) and image understanding tasks to 200:1. We ensure that each batch includes at least one sample from each of the T2I and VLM datasets. The model is optimized with AdamW~\cite{adam} at a learning rate of 2e-5, without warmup.

In the Character-GRPO stage, we utilize the Group Relative Policy Optimization (GRPO) algorithm adapted for flow-matching models. For each input prompt, we sample a group of $G=8$ images to calculate the relative rewards. We employ a relatively conservative optimization strategy with a learning rate of 1e-5 and a batch size of 6. Notably, we set the KL divergence coefficient $\beta$ to 0, relying instead on a strict clipping mechanism to maintain training stability. The clipping range parameters, $\epsilon_{lt}$ and $\epsilon_{gt}$, are both set to $1 \times 10^{-5}$, ensuring that the policy updates remain within a very narrow trust region. The training is conducted using BFloat16 mixed precision to balance numerical stability and computational efficiency.

For the BAGEL model, we use different sampling step counts for training and evaluation: $N_{train}=15$ steps for efficient reward computation during the GRPO loop, and $N_{eval}=50$ steps for high-quality final synthesis. We apply a guidance scale of 4.0. To enhance the stochasticity and structural preservation during personalization, we incorporate Stochastic Differential Equation (SDE) sampling with a noise level of 1.3. We conducted full-parameter tuning, setting the SDE window size to 3. The SDE window range is restricted to the first half of the denoising process ($[0, \lfloor N/2 \rfloor]$) to maintain global structure while allowing for fine-grained detail refinement.

The reward signal $R$ is a multi-objective weighted sum designed to balance aesthetic quality, text alignment, and identity preservation. The components are defined as follows:
\begin{itemize}
\item \textbf{CLIP Similarity Reward ($w=0.45$):} Measured by CLIP-T score to ensure the generated image matches the prompt.
\item \textbf{VQA Consistency Reward ($w=0.3$):} Assessed using a VQA-based scorer.
\item \textbf{Perceptual Diversity Reward ($w=0.1$):} Measured via LPIPS distance within the group of $G$ images to prevent mode collapse.
\item \textbf{Trainset Similarity Penalty ($w=0.15$):} We utilize a DINO-based penalty to prevent overfitting. Specifically, if the DINO similarity exceeds a high threshold (0.9) or falls below a low threshold (0.5), a penalty is applied. This encourages the model to maintain the target identity without generating near-identical replicas or losing the character's essence.
\end{itemize}

\subsection{LLM-as-Judge}
\label{subsec:supp-llm-as-judge}
For the \textbf{Text-based Role Play} task, we employ an ``LLM-as-Judge" methodology for comparison. Specifically, we annotate a specialized set of test questions focused on ``Memorization" and ``Personality." Memorization refers to the model's ability to accurately and comprehensively recall information pertinent to the character it is portraying, including associated people, events, and objects. Personality assesses the model's capacity to emulate the character's thought processes and speech patterns, including linguistic style, tone, and emotional expression across various contexts. However, we observe from training without extended dialogues that models tend to replicate template-like responses from given examples. Although such outputs may align with the character’s Personality, this is not an ideal scenario. To quantify this capability, we provide the LLM with 20 sets of user inputs and model responses, asking it to assign a Diversity score that reflects the model’s ability to produce varied, non-repetitive replies.

Meanwhile, we use the Qwen3~\cite{qwen3} LLM to evaluate the model's role-play text outputs on the Personality, Memorization, and Diversity metrics.  

For Personality and Memorization, each test question corresponds to a User Input–Model Response pair, yielding an individual score; the final score for each metric is the average across all test questions.  

For Diversity, the 20 test questions' User Input–Model Response pairs are combined into a more comprehensive set of model–user interactions, which are jointly evaluated to produce a single Diversity score.

All three scores range from 1 to 7.

Following Character-llm~\cite{characterllm}, our evaluation prompt is displayed in~\cref{tab:personality-metrics,tab:memorization-metrics,tab:diversity-metrics}

\begin{table}[!htbp]
\centering
\caption{Prompt for Personality Metrics.}
\scriptsize
\begin{minipage}{\linewidth}
\textbf{Prompt for Personality Metrics}

\rule{\linewidth}{0.4pt} 

You will be given responses written by an AI assistant mimicking the character [Character Name]. Your task is to rate the performance of [Character Name] using the specific criterion by following the evaluation steps. Below is the data:

\medskip
    
***

Profile:

[Character Profile]

Example Dialogues:

[Example Dialogues]

\medskip

***

Interactions:

[Interactions]  

\medskip

***

Evaluation Criterion:

Personality (1-7): Is the response reflects the personalities and preferences of the character?

\medskip

[Evaluation Steps]  

1. Read through the profile, background, example dialogues, and write the personalities and preferences of the real character.  

2. Read through the interactions and identify the personalities and preferences of the AI assistant.  

3. After having a clear understanding of the interactions, compare the responses to the profile. Look for any consistencies or inconsistencies. Do the responses reflect the character's personalities and preferences?  

4. Use the given scale from 1-7 to rate how well the response reflects the personalities and preferences of the character. 1 being not at all reflective of the character's personalities, and 7 being perfectly reflective of the character's personalities.

\medskip

***

\medskip

First, write out in a step by step manner your reasoning about the criterion to be sure that your conclusion is correct. Avoid simply stating the correct answers at the outset. Then print the score on its own line corresponding to the correct answer. At the end, repeat just the selected score again by itself on a new line.

\rule{\linewidth}{0.4pt} 
\end{minipage}
\label{tab:personality-metrics}
\end{table}

\begin{table}[!htbp]
\centering
\caption{Prompt for Memorization Metrics.}
\scriptsize
\begin{minipage}{\linewidth}
\textbf{Prompt for Memorization Metrics}

\rule{\linewidth}{0.4pt} 

You will be given responses written by an AI assistant mimicking the character [Character Name]. Your task is to rate the performance of [Character Name] using the specific criterion by following the evaluation steps. Below is the data:

\medskip
    
***

Profile:

[Character Profile]

\medskip

***

Interactions:

[Interactions]  

\medskip

***

Evaluation Criterion:

Factual Correctness (1-7): Is the response provides truthful and detailed facts about the character?

\medskip

[Evaluation Steps]  

1. Read through the profile and interaction and identify the key points related to the character.  

2. Read through the responses of the AI assistant and compare them to the profile. Check if the responses are consistent with the character's profile, background, and known facts about the character. 

3. Check whether the responses provide detailed facts about the character or if they are generic responses that could apply to any character. Detailed responses are more factual and contribute positively to the score.  

4. Rate the performance of the AI on a scale of 1-7 for factual correctness, where 1 is the lowest and 7 is the highest based on the Evaluation Criteria.

\medskip

***

\medskip

First, write out in a step by step manner your reasoning about the criterion to be sure that your conclusion is correct. Avoid simply stating the correct answers at the outset. Then print the score on its own line corresponding to the correct answer. At the end, repeat just the selected score again by itself on a new line.

\rule{\linewidth}{0.4pt} 
\end{minipage}
\label{tab:memorization-metrics}
\end{table}

\begin{table}[!htbp]
\centering
\caption{Prompt for Diversity Metrics.}
\scriptsize
\begin{minipage}{\linewidth}
\textbf{Prompt for Diversity Metrics}

\rule{\linewidth}{0.4pt} 

You will be given multiple responses written by an AI assistant to different user inputs. Your task is to rate the performance of the AI assistant on the diversity of its responses by following the evaluation steps. Below is the data:

\medskip

***

Interactions
[Interactions]  

\medskip

***

[Evaluation Criterion]  

Diversity (1-7): How varied and non-repetitive are the AI's responses across different interactions?

\medskip

[Evaluation Steps]  

1. Read through all the provided user inputs and the corresponding model responses.  

2. Analyze the language, structure, and content of the model's responses across the different interactions. Identify any recurring phrases, sentence structures, or response templates.  

3. Assess the degree of repetition. Does the assistant provide unique, context-specific answers for each input, or does it rely on a limited set of formulaic responses?  

4. Use the given scale from 1-7 to rate the diversity of the responses. 1 means the responses are highly repetitive and templated, while 7 means the responses are highly varied, creative, and tailored to each specific user input.

\medskip

***

First, write out in a step by step manner your reasoning about the criterion to be sure that your conclusion is correct. Avoid simply stating the correct answers at the outset. Then print the score on its own line corresponding to the correct answer. At the end, repeat just the selected score again by itself on a new line.

\rule{\linewidth}{0.4pt} 
\end{minipage}
\label{tab:diversity-metrics}
\end{table}

\stopcontents[appendix]
\clearpage